\documentclass{ieeeaccess}
\usepackage{cite}
\usepackage{amsmath,amssymb,amsfonts}
\usepackage{algorithmic}
\usepackage{graphicx}
\usepackage{multirow}
\usepackage{textcomp}
\usepackage{pifont}
\usepackage{hyperref}
\usepackage{bm}
\makeatletter
\AtBeginDocument{\DeclareMathVersion{bold}
\SetSymbolFont{operators}{bold}{T1}{times}{b}{n}
\SetSymbolFont{NewLetters}{bold}{T1}{times}{b}{it}
\SetMathAlphabet{\mathrm}{bold}{T1}{times}{b}{n}
\SetMathAlphabet{\mathit}{bold}{T1}{times}{b}{it}
\SetMathAlphabet{\mathbf}{bold}{T1}{times}{b}{n}
\SetMathAlphabet{\mathtt}{bold}{OT1}{pcr}{b}{n}
\SetSymbolFont{symbols}{bold}{OMS}{cmsy}{b}{n}
\renewcommand\boldmath{\@nomath\boldmath\mathversion{bold}}}
\makeatother

\def\BibTeX{{\rm B\kern-.05em{\sc i\kern-.025em b}\kern-.08em
    T\kern-.1667em\lower.7ex\hbox{E}\kern-.125emX}}

\newcommand{\tick}{\textcolor{green}{\ding{51}}}
\newcommand{\cross}{\textcolor{red}{\ding{55}}}
\begin{document}
%\history{Date of publication xxxx 00, 0000, date of current version xxxx 00, 0000.}
\doi{}

\title{A Survey of Adversarial Defenses in Vision-based Systems: Categorization, Methods and Challenges}
%\author{\uppercase{First A. Author}\authorrefmark{1}, \IEEEmembership{Fellow, IEEE},
%\uppercase{Second B. Author\authorrefmark{2}, and Third C. Author, Jr}.\authorrefmark{3},
%\IEEEmembership{Member, IEEE}}
%\address[1]{National Institute of Standards and Technology, Boulder, CO 80305 USA (e-mail: author@boulder.nist.gov)}
%\address[2]{Department of Physics, Colorado State University, Fort Collins, CO 80523 USA (e-mail: author@lamar.colostate.edu)}
%\address[3]{Electrical Engineering Department, University of Colorado, Boulder, CO 80309 USA}

\author{Nandish Chattopadhyay$^{1}$, Abdul Basit $^{1}$, 
Bassem Ouni$^2$, Muhammad Shafique$^1$ \\
$^1$ eBrain Lab, Division of Engineering, New York University (NYU) Abu Dhabi, UAE \\ $^2$ Technology Innovation Institute (TII), Abu Dhabi, UAE\\
}
\tfootnote{This research was partially funded by Technology Innovation Institute (TII) under the "CASTLE: Cross-Layer Security for Machine Learning Systems IoT" project.}

\corresp{Corresponding author: Nandish Chattopadhyay (e-mail: nc3397@nyu.edu).}

\begin{abstract}
Adversarial attacks have emerged as a major challenge to the trustworthy deployment of machine learning models, particularly in computer vision applications. These attacks have a varied level of potency and can be implemented in both white box and black box approaches. Practical attacks include methods to manipulate the physical world and enforce adversarial behaviour by the corresponding target neural network models. Multiple different approaches to mitigate different kinds of such attacks are available in the literature, each with their own advantages and limitations. In this survey, we present a comprehensive systematization of knowledge on adversarial defenses, focusing on two key computer vision tasks: image classification and object detection. We review the state-of-the-art adversarial defense techniques and categorize them for easier comparison. In addition, we provide a schematic representation of these categories within the context of the overall machine learning pipeline, facilitating clearer understanding and benchmarking of defenses. Furthermore, we map these defenses to the types of adversarial attacks and datasets where they are most effective, offering practical insights for researchers and practitioners. This study is necessary for understanding the scope of how the available defenses are able to address the adversarial threats, and their shortcomings as well, which is necessary for driving the research in this area in the most appropriate direction, with the aim of building trustworthy AI systems for regular practical use-cases. 
\end{abstract}


\titlepgskip=-25pt

\maketitle
\section{Introduction}
The rapid growth of deep learning models has revolutionized numerous industries, with advancements in architecture and training techniques driving their widespread adoption. These models are now integral to practical applications across various fields, including natural language processing, computer vision, autonomous vehicles, and healthcare. Their ability to analyze large datasets, recognize patterns, and make predictions has led to significant breakthroughs in automation, personalized services, and decision-making processes. As deep learning continues to evolve, its impact on both everyday life and specialized industries is expected to grow even further.
Despite the rapid advancements in machine learning (ML) and artificial intelligence, adversarial attacks remain a significant challenge to their widespread adoption in practical applications. The rapid progress of deep learning architectures, exemplified by models like ChatGPT, has demonstrated the widespread applicability of neural networks in everyday life. Substantial efforts and resources have been invested in developing high-performance models, continually enhancing their capabilities. These models have demonstrated their potential in a wide range of applications, including natural language processing (BERT) \cite{devlin2019bert}, biological sciences (AlphaFold) \cite{skolnick2021alphafold}, and text generation (GPT-4) \cite{Liu_2023}. The vulnerability of high-performance neural networks was first demonstrated in 2015~\cite{szegedy}, with initial observations emerging in the domain of computer vision, specifically in image processing tasks. Adversaries were able to generate subtly modified test samples that could mislead trained networks into incorrect classifications. These small, structured perturbations, while imperceptible to human annotators, had a profound impact on model performance, leading to numerous misclassifications~\cite{basic}, \cite{cod_1}, \cite{cod_2}, \cite{cod3}.
Shortly thereafter, intriguing properties such as the transferability of adversarial samples—where adversarial examples crafted for one model could also deceive other models—were uncovered~\cite{papernot}. The research in this domain of adversarial attacks in images has led to competitive works in attacks and defences which have both improved over time. The first adversarial attack was studied as an interesting property of neural networks \cite{szegedy}, but over time, different attack mechanisms have been developed like the Fast Gradient Sign method \cite{FSGM}, the Momentum Iterative version of it \cite{MIFSGM}, the Carlini Wagner attack \cite{cw} and the Projected Gradient Descent attack \cite{pgd} and others \cite{kurakin2018adversarial}. Similarly, multiple defence techniques have been proposed over time, to mitigate such attacks. Some prominent ones include the defensive distillation technique \cite{defensive_distillation} and other filtering mechanisms. These findings spurred significant interest among researchers, resulting in a rapid expansion of the field. Consequently, numerous adversarial attacks were developed, accompanied by a wide array of proposed defense mechanisms~\cite{sir_survey}.
Several techniques have also been proposed for backdoor attacks in neural networks, which are a type of attack implanted into the neural networks during training \cite{chattopadhyay2021robustness}. Notable examples include BlackMarks ~\cite{blackmarks},  DeepMarks ~\cite{deepmarks}, DeepSigns ~\cite{deepsigns} etc. Examples of such attack mitigation efforts include DeepInspect ~\cite{deepinspect} and TABOR ~\cite{tabor} which have been successful in defeating them ~\cite{evasion}, ~\cite{stealing}, . Adversarial attacks however, are performed at the inference phase. 
% Create a floating figure at the bottom of the first page
\begin{figure*}
    \centering
    \includegraphics[width=0.7\textwidth]{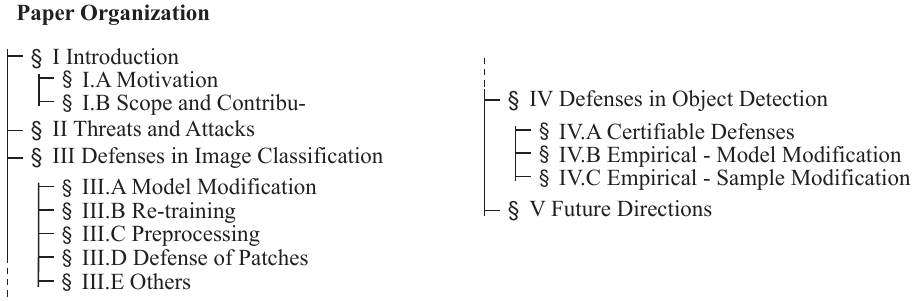} 
    \caption{Overview of the paper organization. This figure illustrates the structure and flow of the sections, highlighting the key topics discussed throughout the paper.}
    \label{fig:example}
\end{figure*}

\subsection{Motivation}
This survey on adversarial defenses is motivated by the increasing recognition of the vulnerability of machine learning models, particularly deep neural networks, to adversarial attacks. Adversarial attacks involve carefully crafted input data that, while often imperceptible to humans, can significantly mislead model predictions. These attacks present a critical threat to the deployment of ML models in real-world applications where security and reliability are paramount. The primary objectives of this survey are as follows: 

\begin{enumerate}
    \item \textbf{Addressing Security Concerns}: As machine learning models are increasingly integrated into critical domains such as autonomous vehicles, healthcare, finance, and security systems, understanding and mitigating their susceptibility to adversarial attacks is imperative. This survey aims to identify the various threats and vulnerabilities these models face and map existing defense techniques to corresponding attack strategies.

    \item \textbf{Evaluating Real-World Impact}: Adversarial attacks can have severe real-world consequences, leading to erroneous predictions and potential security breaches. This survey explores current defense mechanisms and strategies to mitigate these risks, thereby enhancing the reliability of machine learning systems in practical applications.

    \item \textbf{Keeping Up with State-of-the-Art Developments}: Adversarial attacks continue to evolve, with new, sophisticated methods emerging to bypass existing defenses. This survey seeks to provide a comprehensive overview of the latest advancements in adversarial defenses, enabling researchers and practitioners to stay informed about cutting-edge techniques and their implications for model robustness and generalization.

    \item \textbf{Bridging the Gap Between Theory and Practice}: Many adversarial defense strategies remain theoretical, with limited insights into their practical implementation. This survey highlights the effectiveness of various defenses in real-world scenarios, offering guidance to practitioners in selecting and deploying suitable strategies.
\end{enumerate}

By addressing these objectives, this survey contributes to the growing body of knowledge aimed at ensuring the robustness and security of machine learning models against adversarial threats.

% \usepackage{multirow}
% \begin{table*}[]
% \centering 
% \caption{Comparison of most recent surveys in adversarial defenses of vision based tasks.}
% \label{survey}
% \begin{tabular}{lcccccc}
% \hline
% \multicolumn{1}{c}{\multirow{2}{*}{Survey}} & \multirow{2}{*}{Year} & \multirow{2}{*}{\begin{tabular}[c]{@{}c@{}}Number of\\ References\end{tabular}} & \multirow{2}{*}{\begin{tabular}[c]{@{}c@{}}Survey\\ Comparison\end{tabular}} & \multirow{2}{*}{\begin{tabular}[c]{@{}c@{}}Scope (Attacks\\ and Datasets)\end{tabular}} & \multirow{2}{*}{\begin{tabular}[c]{@{}c@{}}Task: Image\\ Classification\end{tabular}} & \multirow{2}{*}{\begin{tabular}[c]{@{}c@{}}Task: Object\\ Detection\end{tabular}} \\
% \multicolumn{1}{c}{} &  &  &  &  &  &  \\ \hline
% Shilin Qiu et al \cite{qiu2019review}& 2019 & 90 & \cross & \cross & \tick & \cross \\
% Qiu et al \cite{qiu2020}& 2020 & 31 & \cross & \cross & \tick & \cross \\
% Akhtar et al  \cite{akhtar2021}& 2021 & 455 & \cross & \tick & \tick & \cross \\
% Khamaiseh et al \cite{khamaiseh2022adversarial}& 2022 & 128 & \cross & \tick & \tick & \cross \\
% Costa et al \cite{costa2023deep}& 2023 & 177 & \cross & \cross & \tick & \cross \\
% Bountakas et al \cite{bountakas2023defense} & 2023 & 127 & \tick & \tick & \tick & \cross \\
% \textbf{Ours} & 2024 & 67 & \tick & \tick & \tick & \tick \\ \hline
% \end{tabular}
% \end{table*}

\begin{table*}[]
\centering 
\caption{Comparison of most recent surveys in adversarial defenses of vision-based tasks.}
\label{survey}
\begin{tabular}{lcccccc}
\hline
\multicolumn{1}{c}{\multirow{2}{*}{Survey (First Author et al.)}} & \multirow{2}{*}{Year} & \multirow{2}{*}{\begin{tabular}[c]{@{}c@{}}Number of\\ References\end{tabular}} & \multirow{2}{*}{\begin{tabular}[c]{@{}c@{}}Survey\\ Comparison\end{tabular}} & \multirow{2}{*}{\begin{tabular}[c]{@{}c@{}}Scope (Datasets,\\ Attacks Covered)\end{tabular}} & \multirow{2}{*}{\begin{tabular}[c]{@{}c@{}}Task: Image\\ Classification\end{tabular}} & \multirow{2}{*}{\begin{tabular}[c]{@{}c@{}}Task: Object\\ Detection\end{tabular}} \\
\multicolumn{1}{c}{} &  &  &  &  &  &  \\ \hline
Chakraborty et al. \cite{chakraborty2018} & 2018 & 79 & \cross & \tick & \tick & \cross \\
Shilin Qiu et al. \cite{qiu2019review} & 2019 & 90 & \cross & \tick & \tick & \tick \\
Qiu et al. \cite{qiu2020} & 2020 & 31 & \cross & \cross & \tick & \cross \\
Silva et al. \cite{silva2020} & 2020 & 136 & \cross & \tick & \tick & \cross \\
Akhtar et al. \cite{akhtar2021} & 2021 & 456 & \cross & \tick & \tick & \cross \\
Khamaiseh et al. \cite{khamaiseh2022adversarial} & 2022 & 128 & \cross & \tick & \tick & \cross \\
Wu et al. \cite{wu2023} & 2023 & 238 & \cross & \tick & \tick & \cross \\
Wang et al. \cite{wang2023} & 2023 & 150 & \cross & \tick & \tick & \cross \\
Bountakas et al. \cite{bountakas2023defense} & 2023 & 127 & \tick & \tick & \tick & \cross \\
Costa et al. \cite{costa2023deep} & 2023 & 177 & \tick & \tick & \tick & \cross \\
\textbf{Ours} & 2024 & 152 & \tick & \tick & \tick & \tick \\ \hline
\end{tabular}
\end{table*}

\subsection{Scope and Contributions}
The primary objective of this survey is to systematically study and organize the state-of-the-art adversarial defense techniques in different vision-based applications to facilitate an easier understanding of the landscape and to identify potential research gaps. Specifically, we focus on two key computer vision tasks: image classification and object detection. While existing surveys extensively cover adversarial attacks and defenses in image classification, a similar level of attention has not been given to object detection—a critical task in many machine learning applications, including autonomous driving and healthcare. The major contributions of this survey are summarized as follows: 
 
\begin{itemize} 
\item Conducting a comprehensive and detailed review of adversarial defenses in both image classification and object detection tasks. 
\item Mapping each defense method to the types of adversarial attacks it mitigates and the datasets where its effectiveness has been demonstrated. 
\item Organizing defense techniques into categories based on common methodologies to highlight their strengths and weaknesses. 
\end{itemize}
In addition, we review existing surveys in this domain and compare their scope and focus with ours, highlighting the unique contributions of this survey. These comparisons are summarized in Table \ref{survey}.

\section{Threats and Attacks}
Let us formally define an adversarial example. To begin with, let us assume that $X$ is the space of input samples. For images as input to the system, $X$ would trivially be the vector space of dimensions equaling the number of pixels in the image. The two classifiers that we are considering here are $f_1$ (sample classifier) and $f_2$ (human annotator). The classifiers have two components each, feature extraction and classification. $X_1$ is the feature space for the sample classifier and $X_2$ is the feature space for the human annotator, where $d_1$ and $d_2$ are norms defined in the spaces $X_1$ and $X_2$ respectively. As shown, $f_1=c_1 \circ g_1$ and $f_2 = c_2 \circ g_2$. %~\cite{theoretical}. 

Let us consider $x \in X$, a training sample. Given $x$, the corresponding adversarial example $x^*$, for a norm $d_2$ defined on the space $X_2$, and a predefined threshold $\delta > 0$, satisfies:
\begin{equation} 
\centering
\label{eq1}
\begin{split}
 f_1(x) \neq f_1(x^*) \quad \mbox{and} \quad f_2(x) = f_2(x^*)\\
 \mbox{such that} \quad d_2(g_2(x),g_2(x^*))< \delta \\  \nonumber 
\end{split}
\end{equation}

\begin{figure}[!htbp]
\centering
\includegraphics[width=\columnwidth]{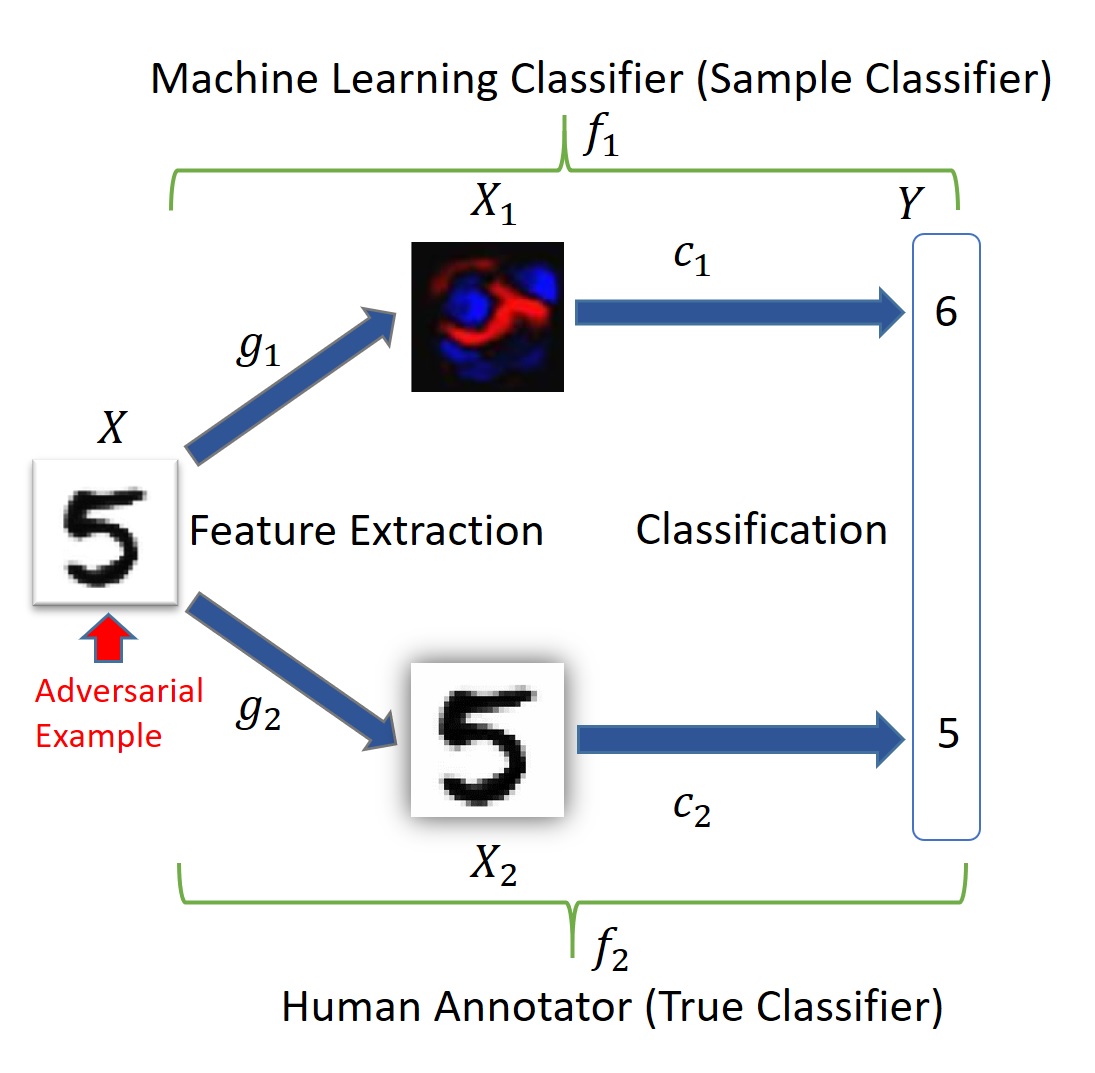}
\label{adv2} 
\caption{An example of adversarial sample exhibiting adversarial behavior.}
\vspace{-10pt}

\end{figure}
Consistent with this definition, there are a variety of different adversarial attacks that are available in the literature, with varying degrees of potency, and targeted at different tasks in computer vision, namely image classification \cite{guesmi2023advrain, googleap, PAE, EOT, lavan, LightAttack, D2P, ACS, projectorattack,ACO, advwatermark, abba, viewfool, slmattack, meta-attack, Invisible-Perturbations, ISP, AdvLB, ACO2, tntattack, pasteattack, advcf, spaa, fakeWeather, advrain}, image recognition \cite{RP2, darts, roguesigns, PSGAN, advcam, OPAD, shadow, physgan}, object detection \cite{guesmi2023advart, guesmi2024dap, tpatch, dpatch, dpatch2, objecthider, lpattack, switchpatch, extendedrp2, shapeshifter, nestedae, translucent-patch, SLAP, adversarialrain, advrd, invisiblecloak, Adversarialyolo, UPC, adversarialtshirt, invisiblecloak2, NAP, LAP, advtexture, advart, PatchOI, dap, bulb, qrattack, hotcold, AIP, AdvIB, camou, ERattack, dynamicpatch, PG} and depth estimation \cite{guesmi2024saam, guesmi2024ssap}. 
As adversarial attacks get more advanced, the recent developments in this direction involve practical physical attacks which are implementable like stickers etc and works in realistic settings \cite{guesmi2023physical}. Table~\ref{adv_attacks} summarizes the most recently published practical adversarial attacks, often referred to as physical attacks. These outline the state-of-the-art threats for machine learning systems.

\begin{table*}[!htp]
\centering
  \caption{Main physical adversarial attack methods in Computer Vision: Form, Application and Venue adapted from \cite{guesmi2023physical}.}
  \label{adv_attacks}
  \begin{tabular}{lccc}
    \hline
       \textbf{Method}  & \textbf{Form}  & \textbf{Application}  & \textbf{Venue}  \\
    \hline 
            GoogleAp \cite{googleap} & Patch & Image Classification & NIPS 2017\\
            PAE \cite{PAE} & Printed images   & Image Classification &  AISS 2018   \\
            EOT \cite{EOT} & 3D-printed object & Image Classification & PMLR 2018\\
            LaVAN \cite{lavan} & Patch & Image Classification & ICML 2018\\
            LightAttack \cite{LightAttack}  & Light based &  Image Classification & AAAI-S 2018 \\
            D2P \cite{D2P} & Printed images & Image Classification & AAAI 2019\\
            ACS \cite{ACS} & Sticker & Image Classification & PMLR 2019\\
            ProjectorAttack \cite{projectorattack} &   Light based & Image Classification & S\&P 2019  \\
            Adversarial ACO  \cite{ACO} & Patch & Image Classification & ECCV 2020\\
            Adv-watermark  \cite{advwatermark} & Patch & Image Classification & ACM MM 2020\\
            ABBA  \cite{abba} & Printed image &  Image Classification &  NeurIPS 2020   \\
            ViewFool \cite{viewfool} &  Position    & Image Classification &  NeurIPS 2020  \\
            SLMAttack  \cite{slmattack}     & Light based &  Image Classification  & ArXiv 2021     \\
            Meta-Attack \cite{meta-attack} & Image & Image Classification &ICCV 2021 \\
            %DAS  & \cite{DAS} & Sticker & Classification & CVPR & 2021\\
            Invisible perturbations \cite{Invisible-Perturbations} & Camera & Image Classification & CVPR 2021\\
            Adversarial ISP  \cite{ISP} & Camera & Image Classification & CVPR 2021\\
            AdvLB \cite{AdvLB} & Light based & Image Classification & CVPR 2021 \\
            Adversarial ACO2 \cite{ACO2} & Patch & Image Classification & IEEE TIP 2022\\ 
            TnT attack\cite{tntattack} & Patch & Image Classification & TIFS 2022 \\
            Copy/Paste Attack  \cite{pasteattack} & Patch &  Image Classification &  NeurIPS 2022 \\
            AdvCF \cite{advcf} & Sticker & Image Classification & Arxiv 2022 \\
            SPAA \cite{spaa}   & Light based &  Image Classification  &  VR 2022 \\ 
            FakeWeather \cite{fakeWeather} & Sticker & Image Classification & IJCNN 2022\\
            AdvRain  \cite{advrain} & Sticker & Image Classification & Arxiv 2023\\
             %%%           
            $RP_2$ \cite{RP2} & Patch & Traffic Sign Detection & CVPR 2018\\
            DARTS \cite{darts} & Image & Traffic Sign Detection  & Arxiv 2018\\
            RogueSigns \cite{roguesigns}& Printed images & Traffic Sign Detection & Arxiv 2018 \\ %%%
            PS-GAN \cite{PSGAN} & Patch & Traffic Sign Detection  &AAAI 2019\\
            AdvCam \cite{advcam} & Image & Traffic Sign Detection  & CVPR 2020 \\
            OPAD \cite{OPAD} & Light based & Traffic Sign Detection & ICCV 2021\\
            Adversarial Shadow \cite{shadow} & Light based &  Traffic Sign Detection & CVPR 2022\\
            %%%
            PhysGAN \cite{physgan} & Image & Steering Model& CVPR 2020\\
            %%%%%%%%%%%%%%%%%%%%%%%%%%%%%%%%%%%%%%%%%%%%%%%%%%%%%%%%%%%%%%%%%%%%%%%%%%%%%%%%
            TPatch \cite{tpatch} & Acoustics & Detection \& classification & Arxiv 2023\\  
            
            DPATCH \cite{dpatch} & Patch & Object Detection & AAAI 2019\\
            Dpatch2 \cite{dpatch2} & Patch & Object Detection & ArXiv 2019  \\
            Object Hider \cite{objecthider} &  Patch  & Object Detection   &   ArXiv 2020\\
            LPAttack \cite{lpattack} & Patch &  Object Detection & AAAI 2020 \\
            SwitchPatch \cite{switchpatch}& Patch & Object Detection & ArXiv 2022\\
            Extended RP2 \cite{extendedrp2} & Patch & Sign Detection & USENIX 2018\\
            ShapeShifter \cite{shapeshifter} & Image & Sign Detection & ECML PKDD 2018\\
            NestedAE \cite{nestedae} & Patch & Sign Detection    &  CCS 2019    \\
            Translucent Patch \cite{translucent-patch} & Sticker & Sign Detection & CVPR 2021\\
            SLAP \cite{SLAP} & Light based & Sign Detection & USENIX 2021\\  
            Adversarial Rain \cite{adversarialrain} & Sticker & Sign Detection & Arxiv 2022\\ 
            AdvRD \cite{advrd} & Sticker & Sign Detection & Arxiv 2023 \\

            Invisible Cloak \cite{invisiblecloak} & Clothing & Person Detection & UEMCON 2018\\  
            Adversarial YOLO  \cite{Adversarialyolo} & Patch & Person Detection & CVPR 2019\\
            UPC \cite{UPC} & Clothing & Person Detection & CVPR 2020\\
            Adversarial T-shirt \cite{adversarialtshirt} & Clothing & Person Detection & ECCV 2020 \\ 
            Invisible Cloak2 \cite{invisiblecloak2} & Clothing & Person Detection & ECCV 2020\\
            NAP \cite{NAP} & Clothing & Person Detection & ICCV 2021\\
            LAP \cite{LAP} & Clothing & Person Detection & ACM MM 2021\\  
            %TC-EGA \cite{TC-EGA} & Clothing  & Person detection & CVPR 2022 \\    
            AdvTexture \cite{advtexture} & Clothing &  Person Detection & CVPR 2022   \\
            AdvART \cite{advart} & Patch & Person Detection & ArXiv 2023 \\  
            Patch of Invisibility \cite{PatchOI}  & Patch & Person Detection & ArXiv 2023 \\
            DAP \cite{dap} & Clothing & Person Detection & ArXiv 2023 \\              
            Adversarial Bulbs \cite{bulb} & Bulb & Infrared Person Detection & AAAI 2021\\
            QRAttack \cite{qrattack} & Clothing & Infrared Person Detection & CVPR 2022\\    
            HOTCOLD \cite{hotcold} & Clothing & Infrared Person Detection & ArXiv 2022\\
            AIP \cite{AIP}  & Clothing & Infrared Person Detection & ArXiv 2023 \\
            AdvIB \cite{AdvIB}  &  Clothing  &   Infrared Person Detection      &  ArXiv 2023 \\
            %MeshAdv  & \cite{meshadv}  &    &         &  CVPR  & 2019 \\
            %FIR/ERG  & \cite{FIR} & Patch & Sign detection & CCS & 2019 \\
            CAMOU \cite{camou} & Sticker & Vehicle Detection & ICLR 2019\\
            ER Attack \cite{ERattack} & Sticker & Vehicle Detection & ArXiv 2020\\
            ScreenAttack  \cite{dynamicpatch} & Patch & Vehicle Detection & ArXiv 2020\\
            PG \cite{PG} & Acoustics & Vehicle Detection & S\&P 2021\\

            %%%%%%%%%%%%%%%%%%%%%%%%%%%%%%%%%%%%%%%%%%%%%%%%%%%%%%%%%%%%%%%%%%%%%%

 \hline
\end{tabular} %}
\end{table*}

\section{Defenses in Image Classification}
\begin{figure}[!htbp]
\centering
\includegraphics[width=\columnwidth]{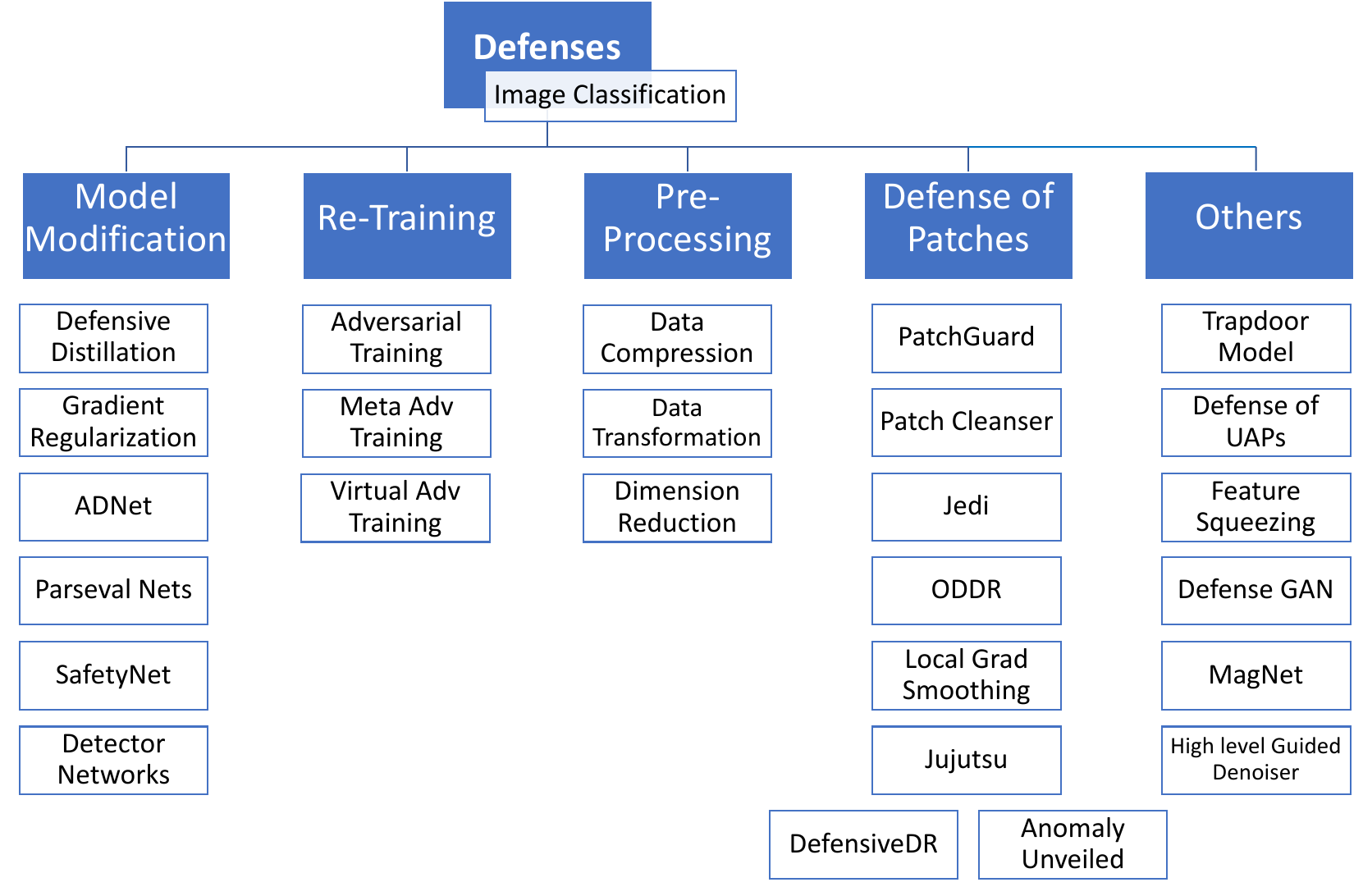}
\caption{Organization of different approaches for defenses against attacks on image classification tasks in vision based systems.  }
\label{classification_image}
\end{figure}

With the rapidly evolving landscape of adversarial attacks on image classification tasks, which are becoming increasingly potent and sophisticated, there is a pressing need to identify, categorize, and evaluate effective adversarial defense mechanisms to mitigate them. It is important to note that specific defenses are often robust against particular types of attacks, and no single mechanism can comprehensively address all adversarial threats.
To provide clarity, we have systematically organized and tabulated the existing literature, highlighting the applicability of defenses and their robustness against specific attack types. This information is presented in Table~\ref{classification_table_basic} and Table~\ref{classification_table_patch}. Furthermore, for a clearer understanding, we have grouped the defense mechanisms based on the similarity in their approaches, as illustrated in Figure~\ref{classification_image}.
The following sections introduce each of these defense categories, briefly describing their key characteristics, followed by a detailed explanation of the individual methods.

\subsection{Model Modification}
Adversarial defenses for image classification often involve modifying the model architecture or the training process to improve robustness against adversarial attacks. Several strategies have been proposed in the literature, including:
\begin{itemize}
    \item Defensive Distillation \cite{defensive_distillation}.
    \item Gradient Regularization \cite{Gradient_regularization}.
    \item ADNet \cite{adnet}.
    \item Parseval Nets \cite{parseval}.
    \item SafetyNet \cite{safetynet}.
    \item Detector Sub-Networks \cite{detector_net}.
\end{itemize}

These approaches focus on enhancing the model’s resilience to adversarial perturbations, ensuring more reliable and secure performance in real-world scenarios. Each of these techniques will be discussed in detail in the subsequent sections.

\begin{table*}[!htbp]
\centering
\caption{Adversarial defenses in image classification tasks for imperceptible attacks.}
\label{classification_table_basic}
\begin{tabular}{llll}
\hline
\multirow{2}{*}{\textbf{Defense}} & \multirow{2}{*}{\textbf{Attacks Covered}} & \multirow{2}{*}{\textbf{Datasets Used}} & \multirow{2}{*}{\textbf{Best Performance}}  \\
 &  &  &  \\ \hline
\multirow{2}{*}{\begin{tabular}[c]{@{}l@{}}Defensive\\ Distillation \cite{defensive_distillation}\end{tabular}} & \multirow{2}{*}{FGSM \cite{FSGM}, JSMA \cite{jsma}} & \multirow{2}{*}{MNIST \cite{mnist}, CIFAR-10 \cite{cifar}} & \multirow{2}{*}{95.8\% (MNIST)}\\
 &  &  &  \\
\multirow{2}{*}{\begin{tabular}[c]{@{}l@{}}Gradient\\ Regularization \cite{grad_reg}\end{tabular}} & \multirow{2}{*}{FGSM \cite{FSGM}, JSMA \cite{jsma}} & \multirow{2}{*}{\begin{tabular}[c]{@{}l@{}}MNIST \cite{mnist}, CIFAR-10 \cite{cifar},\\ ImageNet \cite{imagenet}\end{tabular}}  & \multirow{2}{*}{96.5\% (MNIST)}\\
 &  &  &  \\
\multirow{2}{*}{ADNet \cite{adnet}} & \multirow{2}{*}{\begin{tabular}[c]{@{}l@{}}FSGM \cite{FSGM}, BIM \cite{bim},\\ DeepFool \cite{deepfool}, C\&W \cite{cw}\end{tabular}} & \multirow{2}{*}{ImageNet \cite{imagenet}}  & \multirow{2}{*}{93.84\% (ImageNet)}\\
 &  &  &  \\
\multirow{2}{*}{\begin{tabular}[c]{@{}l@{}}Parseval\\ Networks \cite{parseval}\end{tabular}} & \multirow{2}{*}{FGSM \cite{FSGM}} & \multirow{2}{*}{CIFAR-10 \cite{cifar}, CIFAR-100} & \multirow{2}{*}{96.28\% (CIFAR-10)} \\
 &  &  &  \\
\multirow{2}{*}{\begin{tabular}[c]{@{}l@{}}Safety\\ Nets \cite{safetynet}\end{tabular}} & \multirow{2}{*}{\begin{tabular}[c]{@{}l@{}}FGSM \cite{FSGM}, L-BFGS,\\ DeepFool \cite{deepfool}\end{tabular}} & \multirow{2}{*}{CIFAR-10 \cite{cifar}, ImageNet \cite{imagenet}} & \multirow{2}{*}{Rejects Adversarial samples} \\
 &  &  &  \\
\multirow{2}{*}{\begin{tabular}[c]{@{}l@{}}Detector\\ Networks \cite{detector_net}\end{tabular}} & \multirow{2}{*}{\begin{tabular}[c]{@{}l@{}}FGSM \cite{FSGM}, DeepFool \cite{deepfool},\\ BIM \cite{bim}\end{tabular}} & \multirow{2}{*}{MNIST \cite{mnist}, CIFAR-10 \cite{cifar}} & \multirow{2}{*}{Rejects Adversarial samples} \\
 &  &  &  \\
\multirow{2}{*}{\begin{tabular}[c]{@{}l@{}}Adversarial\\ Training \cite{adv_train}\end{tabular}} & \multirow{2}{*}{FGSM \cite{FSGM}, JSMA \cite{jsma}} & \multirow{2}{*}{CIFAR-10 \cite{cifar}, ImageNet \cite{imagenet}} & \multirow{2}{*}{83.5\% (SVHN)} \\
 &  &  &  \\
\multirow{2}{*}{\begin{tabular}[c]{@{}l@{}}Meta Adversarial\\ Training \cite{meta}\end{tabular}} & \multirow{2}{*}{FGSM \cite{FSGM}} & \multirow{2}{*}{Tiny ImageNet} & \multirow{2}{*}{59\% (Tiny ImageNet)}\\
 &  &  &  \\
\multirow{2}{*}{\begin{tabular}[c]{@{}l@{}}Data\\ Compression \cite{comdefend}\end{tabular}} & \multirow{2}{*}{FGSM \cite{FSGM}} & \multirow{2}{*}{CIFAR-10 \cite{cifar}} & \multirow{2}{*}{89\% (CIFAR-10)}\\
 &  &  &  \\
\multirow{2}{*}{\begin{tabular}[c]{@{}l@{}}Data\\ Transformation \cite{feature_distillation}\end{tabular}} & \multirow{2}{*}{\begin{tabular}[c]{@{}l@{}}FGSM \cite{FSGM}, C\&W \cite{cw},\\ BPDA, DeepFool \cite{deepfool}\end{tabular}} & \multirow{2}{*}{ImageNet \cite{imagenet}} & \multirow{2}{*}{99\% (MNIST)} \\
 &  &  &  \\
\multirow{2}{*}{\begin{tabular}[c]{@{}l@{}}Dimension\\ Reduction \cite{chattopadhyay, chattopadhyay2021}\end{tabular}} & \multirow{2}{*}{\begin{tabular}[c]{@{}l@{}}FGSM \cite{FSGM}, PGD \cite{pgd},\\ MI-FGSM \cite{MIFSGM}\end{tabular}} & \multirow{2}{*}{\begin{tabular}[c]{@{}l@{}}MNIST \cite{mnist}, CIFAR-10 \cite{cifar},\\ ImageNet \cite{imagenet}\end{tabular}} & \multirow{2}{*}{98.6\% (MNIST)} \\
 &  &  &  \\
\multirow{2}{*}{\begin{tabular}[c]{@{}l@{}}Trapdoor based\\ defense \cite{trapdoor}\end{tabular}} & \multirow{2}{*}{\begin{tabular}[c]{@{}l@{}}C\&W \cite{cw}, ElasticNet, PGD \cite{pgd},\\ BPDA, SPSA, FGSM \cite{FSGM}\end{tabular}} & \multirow{2}{*}{\begin{tabular}[c]{@{}l@{}}MNIST \cite{mnist}, GTSRB,\\ CIFAR-10 \cite{cifar}\end{tabular}} & \multirow{2}{*}{Rejects Adversarial samples}  \\
 &  &  &  \\
\multirow{2}{*}{\begin{tabular}[c]{@{}l@{}}Defense against\\ UAPs \cite{UAP}\end{tabular}} & \multirow{2}{*}{\begin{tabular}[c]{@{}l@{}}FGSM \cite{FSGM}, PGD \cite{pgd},\\ DeepFool \cite{deepfool}, F-UAP\end{tabular}} & \multirow{2}{*}{CIFAR-10 \cite{cifar}, ImageNet \cite{imagenet}} & \multirow{2}{*}{64\% (MNIST)} \\
 &  &  &  \\
\multirow{2}{*}{\begin{tabular}[c]{@{}l@{}}Feature\\ Squeezing \cite{feature_squeezing}\end{tabular}} & \multirow{2}{*}{\begin{tabular}[c]{@{}l@{}}FGSM \cite{FSGM}, DeepFool \cite{deepfool},\\ BIM \cite{bim}, JSMA \cite{jsma}\end{tabular}} & \multirow{2}{*}{\begin{tabular}[c]{@{}l@{}}MNIST \cite{mnist}, CIFAR-10 \cite{cifar},\\ ImageNet \cite{imagenet}\end{tabular}} & \multirow{2}{*}{97\% (MNIST)} \\
 &  &  &  \\
\multirow{2}{*}{DefenseGAN \cite{defense_gan}}  & \multirow{2}{*}{\begin{tabular}[c]{@{}l@{}}FGSM \cite{FSGM},\\ C\&W \cite{cw}\end{tabular}} & \multirow{2}{*}{\begin{tabular}[c]{@{}l@{}}MNIST \cite{mnist},\\ Fashion-MNIST\end{tabular}} & \multirow{2}{*}{49.3\% (CIFAR-10)} \\
 &  &  &  \\
\multirow{2}{*}{MagNet \cite{magnet}} & \multirow{2}{*}{\begin{tabular}[c]{@{}l@{}}FGSM \cite{FSGM}, IFGSM,\\ C\&W \cite{cw}, DeepFool \cite{deepfool}\end{tabular}} & \multirow{2}{*}{\begin{tabular}[c]{@{}l@{}}MNIST \cite{mnist},\\ CIFAR-10 \cite{cifar}\end{tabular}} & \multirow{2}{*}{94\% (MNIST)} \\
 &  &  &  \\
\multirow{2}{*}{\begin{tabular}[c]{@{}l@{}}High level\\ Guided Denoiser \cite{hgd}\end{tabular}} & \multirow{2}{*}{\begin{tabular}[c]{@{}l@{}}FGSM \cite{FSGM}, IFGSM \cite{MIFSGM},\\ PGD \cite{pgd}\end{tabular}} & \multirow{2}{*}{ImageNet \cite{imagenet}} & \multirow{2}{*}{72.2\% (ImageNet)} \\
 &  &  &  \\ \hline
\end{tabular}
\end{table*}

\begin{table*}[!htbp]
\centering
\caption{Adversarial defenses in image classification tasks for patch based attacks.}
\label{classification_table_patch}
\begin{tabular}{llll}
\hline
\multirow{2}{*}{\textbf{Defense}} & \multirow{2}{*}{\textbf{Attacks Covered}} & \multirow{2}{*}{\textbf{Datasets Used}} & \multirow{2}{*}{\textbf{Best Performance}}  \\
 &  &  &  \\ \hline
\multirow{2}{*}{\begin{tabular}[c]{@{}l@{}}Patch\\ Guard \cite{patchguard}\end{tabular}} & \multirow{2}{*}{\begin{tabular}[c]{@{}l@{}}Patch\\ (LaVAN \cite{lavan}, Google AP \cite{googleap})\end{tabular}} & \multirow{2}{*}{\begin{tabular}[c]{@{}l@{}}ImageNet \cite{imagenet}, ImageNette,\\ CIFAR-10 \cite{cifar}\end{tabular}} & \multirow{2}{*}{92.9\% (ImageNette)} \\
 &  &  &  \\
\multirow{2}{*}{\begin{tabular}[c]{@{}l@{}}Patch\\ Cleanser \cite{patchcleanser}\end{tabular}} & \multirow{2}{*}{\begin{tabular}[c]{@{}l@{}}Patch\\ (LaVAN \cite{lavan}, Google AP \cite{googleap})\end{tabular}} & \multirow{2}{*}{\begin{tabular}[c]{@{}l@{}}ImageNet \cite{imagenet}, ImageNette,\\ CIFAR-10 \cite{cifar}\end{tabular}} & \multirow{2}{*}{97.5\% (ImageNette) } \\
 &  &  &  \\
\multirow{2}{*}{Jedi \cite{jedi}} & \multirow{2}{*}{\begin{tabular}[c]{@{}l@{}}Patch\\ (LaVAN \cite{lavan}, Google AP \cite{googleap})\end{tabular}} & \multirow{2}{*}{ImageNet \cite{imagenet}} & \multirow{2}{*}{64.3\% (ImageNet)} \\
 &  &  &  \\
\multirow{2}{*}{\begin{tabular}[c]{@{}l@{}}Outlier Detection \&\\ Dimension Reduction \cite{oddr}\end{tabular}} & \multirow{2}{*}{\begin{tabular}[c]{@{}l@{}}Patch\\ (LaVAN \cite{lavan}, Google AP \cite{googleap})\end{tabular}} & \multirow{2}{*}{ImageNet \cite{imagenet}, CalTech 101 \cite{caltech}} & \multirow{2}{*}{91.1\% (CalTech-101)} \\
 &  &  &  \\
\multirow{2}{*}{\begin{tabular}[c]{@{}l@{}}Local Gradient\\ Smoothing \cite{lgs}\end{tabular}} & \multirow{2}{*}{Patch (LaVAN \cite{lavan})} & \multirow{2}{*}{ImageNet \cite{imagenet}} & \multirow{2}{*}{70.90\% (ImageNet)} \\
 &  &  &  \\
\multirow{2}{*}{Jujutsu \cite{jujutsu}} & \multirow{2}{*}{Patch (LaVAN \cite{lavan})} & \multirow{2}{*}{\begin{tabular}[c]{@{}l@{}}ImageNet \cite{imagenet}, ImageNette,\\ CelebA, Place365\end{tabular}} & \multirow{2}{*}{77.5\% (ImageNet)} \\
 &  &  &  \\ 
\multirow{2}{*}{DefensiveDR \cite{defensivedr}} & \multirow{2}{*}{Patch (LaVAN \cite{lavan}, Google AP \cite{googleap})} & \multirow{2}{*}{\begin{tabular}[c]{@{}l@{}}ImageNet \cite{imagenet}\\ CalTech-101 \cite{caltech}\end{tabular}} & \multirow{2}{*}{66.2\% (ImageNet)}  \\
 &  &  &  \\
\multirow{2}{*}{\begin{tabular}[c]{@{}l@{}}Anomaly \\ Unveiled \cite{anomalyunveiled}\end{tabular}} & \multirow{2}{*}{\begin{tabular}[c]{@{}l@{}}Patch\\ (LaVAN \cite{lavan}, Google AP \cite{googleap})\end{tabular}} & \multirow{2}{*}{ImageNet \cite{imagenet}, CalTech 101 \cite{caltech}} & \multirow{2}{*}{67.1\% (ImageNet)} \\
 &  &  &  \\ \hline
\end{tabular}
\end{table*}

\begin{figure*}[!htbp]
\centering
\includegraphics[width=0.9\textwidth]{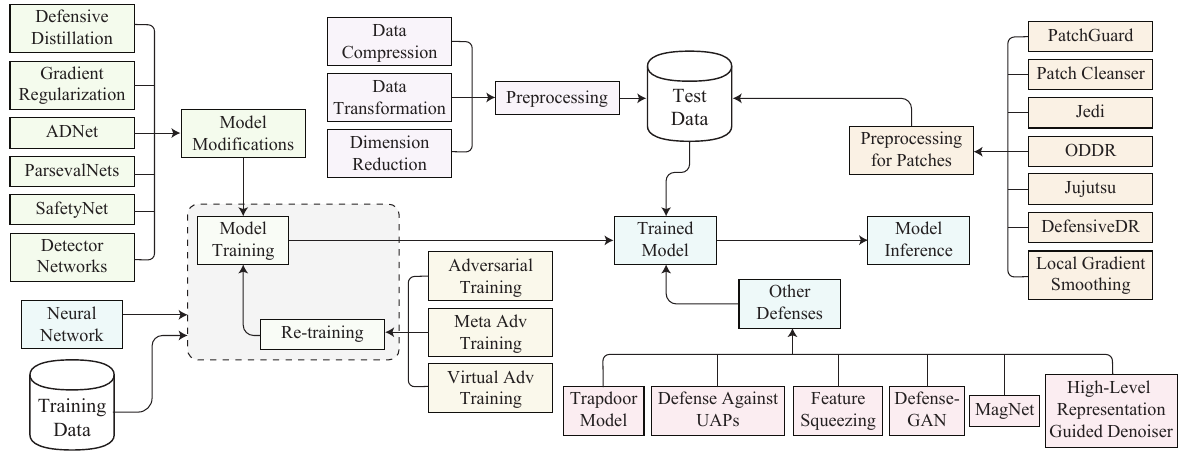}
\caption{Schematic representation of the integration of the defense techniques in different parts of a standard machine learning pipeline for image classification tasks.  }
\label{classification_image_org}
\end{figure*}

\subsubsection{Defensive Distillation}

\begin{figure}[!htbp]
\centering
\includegraphics[width=\columnwidth]{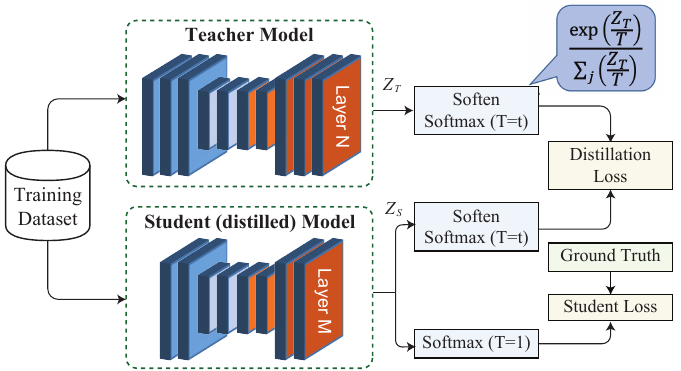}
\caption{Defensive distillation methodology overview: Knowledge Distillation (KD) trains a student model to mimic a teacher model, reducing sensitivity to input perturbations. While effective against white-box attacks, KD remains vulnerable to black-box attacks. The proposed approach improves robustness by training the student to learn a different latent space instead of mimicking the teacher's outputs.}
\label{defence_categories} 
\end{figure}

In various classical machine learning problems, deep learning algorithms have exhibited exceptional performance. However, recent research has indicated that, akin to other machine learning techniques, deep learning is susceptible to adversarial samples—inputs deliberately designed to manipulate a deep neural network (DNN) into producing predetermined outputs chosen by adversaries. Such attacks pose a significant threat to the security of systems relying on DNNs, potentially leading to severe consequences. Instances include the possibility of autonomous vehicle collisions, the evasion of content filters by illicit or illegal content, and manipulation of biometric authentication systems for unauthorized access.

This study introduces a defensive mechanism named defensive distillation \cite{defensive_distillation}, aiming to diminish the impact of adversarial samples on DNNs. The research involves an analytical exploration of the generalizability and robustness properties achieved by employing defensive distillation during DNN training. Additionally, the study empirically assesses the effectiveness of these defense mechanisms in adversarial scenarios involving two DNNs. The results demonstrate that defensive distillation can substantially decrease the efficacy of adversarial sample creation, reducing it from 95\% to less than 0.5\% for a specific DNN under investigation. This remarkable improvement can be attributed to the fact that distillation causes the gradients utilized in the creation of adversarial samples to decrease by a factor of $10^{30}$. Furthermore, the study reveals that distillation increases the average minimum number of features that must be modified to generate adversarial samples by approximately 800\% in one of the tested DNNs.

\subsubsection{Gradient Regularization}
Despite their effectiveness in a range of computer vision tasks, deep neural networks (DNNs) face susceptibility to adversarial attacks, thereby limiting their utility in security-critical systems. Recent studies have unveiled the feasibility of creating imperceptibly altered image inputs, known as adversarial examples, capable of deceiving well-trained DNN classifiers into making arbitrary predictions. In response to this challenge, a proposed training approach, labeled "deep defense," introduces a core concept of incorporating an adversarial perturbation-based regularizer into the classification objective. This integration enables the trained models to actively and precisely resist potential attacks. The optimization process for the entire problem mirrors the training of a recursive network \cite{grad_reg}.

Empirical findings highlight the superior performance of the proposed method compared to training methodologies involving adversarial/Parseval regularizations. This holds true across various datasets, including MNIST, CIFAR-10, and ImageNet, as well as with different DNN architectures, indicating the robustness and efficacy of the "deep defense" strategy.

\subsubsection{ADNet}
The method's detection strategy relies on the adversarial detection network (ADNet) \cite{adnet}, which acquires its detection capabilities through a hierarchical learning process from input images. During this process, the input images traverse through both convolutional and composite layers.

An advantageous attribute of the ADNet technique is its ability to deceive deep models during the test phase. Operating as an independent module, it can identify adversarial examples without being tied to a specific model. Furthermore, it can function as a discreet element within a broader intelligent system. This inherent strength renders the ADNet resilient against attacks directed at itself. This contrasts with many existing decision networks that depend on the internal states of a network during the test phase, exposing them to potential attackers. Additionally, it is noteworthy that these conventional methods are incapable of handling pixel-level attacks.

\subsubsection{Parseval}
This work presents Parseval networks \cite{parseval}, a variant of deep neural networks characterized by the constraint that the Lipschitz constant of linear, convolutional, and aggregation layers is kept smaller than 1. The motivation behind Parseval networks stems from both empirical and theoretical analyses, specifically examining the robustness of predictions made by deep neural networks when subjected to adversarial perturbations in their input.

A key attribute of Parseval networks is the maintenance of weight matrices in linear and convolutional layers as (approximately) Parseval tight frames, extending the concept of orthogonal matrices to non-square matrices. The description includes details on how these constraints can be efficiently upheld during stochastic gradient descent (SGD). Parseval networks demonstrate competitive accuracy on CIFAR-10/100 and Street View House Numbers (SVHN), outperforming their vanilla counterparts in robustness against adversarial examples. Additionally, Parseval networks exhibit tendencies to train faster and make more efficient use of the networks' full capacity, contributing to their overall effectiveness.

\subsubsection{SafetyNet}
The method presented outlines the creation of a network that poses significant challenges for generating adversarial samples, particularly for existing techniques like DeepFool. Through this construction, valuable insights into the functioning of deep networks are revealed. The authors provide a thorough analysis suggesting the robustness of their approach, supported by experimental evidence demonstrating its resilience against both Type I and Type II attacks across various standard networks and datasets.

This SafetyNet architecture \cite{safetynet} finds application in the innovative SceneProof system, designed for reliably determining whether an image depicts a genuine scene. SceneProof is tailored for images accompanied by depth maps (RGBD images) and assesses the consistency of image-depth map pairs. Its effectiveness relies on the inherent difficulty of generating realistic depth maps for images during post-processing. The study showcases that the SafetyNet remains robust even when confronted with adversarial examples generated through currently known attacking approaches.

\subsubsection{Detector Sub-Network}
Research indicates the susceptibility of deep neural network (DNN) based classifiers to imperceptible adversarial perturbations, leading to incorrect and highly confident predictions. The authors propose an unsupervised learning method designed to identify adversarial inputs without requiring knowledge of the attackers \cite{detector_net}. This approach aims to capture the intrinsic properties of a DNN classifier, specifically focusing on the output distributions of hidden neurons when presented with natural images. Notably, the proposed technique can be easily applied to any DNN classifiers or integrated with other defense strategies to enhance overall robustness. Experimental results affirm that the approach showcases state-of-the-art effectiveness in defending against both black-box and gray-box attacks.

\subsection{Re-Training}
Re-training models with specific samples often turn out to be a mechanism for making the models robust against attacks. During training, adversarial examples are generated and used to augment the training dataset. The model is then trained on this augmented dataset, making it more robust to similar adversarial inputs during inference. Adversarial training has shown success in improving the robustness of models against various adversarial attacks. The most useful ones are described in details below.

\begin{figure}[!htbp]
\centering
\includegraphics[width=0.9\columnwidth]{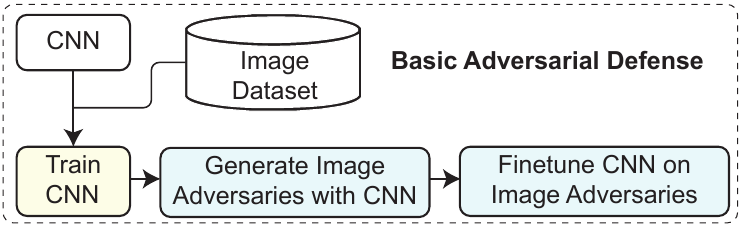}
\caption{Overview of adversarial training.}
\label{adv_training} 
\end{figure}

\subsubsection{Adversarial Training}
%Adversarial examples refer to manipulated inputs intended to deceive machine learning models. Adversarial training incorporates such examples into the training dataset to enhance robustness. To extend this technique to large datasets, perturbations are generated using rapid single-step methods that maximize a linear approximation of the model's loss. This study reveals that this adversarial training form converges to a degenerate global minimum, where small curvature artifacts near data points obscure a linear approximation of the loss. Consequently, the model learns to generate weak perturbations rather than defend against stronger ones. Adversarial training is shown to remain vulnerable to black-box attacks, including the transfer of perturbations from undefended models and a novel single-step attack that escapes the non-smooth vicinity of input data via a small random step. The introduction of Ensemble Adversarial Training, a technique augmenting training data with perturbations transferred from other models, enhances robustness to black-box attacks on ImageNet. However, subsequent research reveals that more sophisticated black-box attacks significantly increase transferability and reduce model accuracy.

Adversarial examples are manipulated inputs designed to mislead machine learning models. Adversarial training integrates such examples into the training dataset to improve robustness. When applied to large datasets, perturbations are generated using fast single-step methods that maximize a linear approximation of the model's loss. This study demonstrates that this form of adversarial training converges to a degenerate global minimum, where minor curvature artifacts near data points distort the linear loss approximation. As a result, the model learns to produce weak perturbations rather than effectively defending against stronger ones. Adversarial training remains susceptible to black-box attacks, including perturbation transfers from undefended models and a novel single-step attack that bypasses the non-smooth region near input data through a small random step. The introduction of Ensemble Adversarial Training, which strengthens training data with perturbations from other models, improves resilience to black-box attacks on ImageNet. However, later research indicates that more advanced black-box attacks significantly enhance transferability and lower model accuracy.

\subsubsection{Meta Adversarial Training}
%Adversarial training is the most effective defense against image-dependent adversarial attacks. However, tailoring adversarial training to universal patches is computationally expensive since the optimal universal patch depends on the model weights which change during training. We propose meta adversarial training (MAT) \cite{meta}, a novel combination of adversarial training with meta-learning, which overcomes this challenge by meta-learning universal patches along with model training. MAT requires little extra computation while continuously adapting a large set of patches to the current model. MAT considerably increases robustness against universal patch attacks on image classification tasks. 

Adversarial training serves as the most effective defense against image-dependent adversarial attacks. However, adapting adversarial training to universal patches is computationally demanding, as the optimal universal patch relies on model weights that change throughout training. Meta Adversarial Training (MAT) \cite{meta} is introduced as a novel approach that integrates adversarial training with meta-learning to address this challenge. By meta-learning universal patches alongside model training, MAT minimizes additional computational costs while continuously adjusting a large set of patches to the evolving model. This technique significantly enhances robustness against universal patch attacks in image classification tasks.

\subsubsection{Stability Adversarial Training}
In this study, the focus is on the algorithmic stability of a generic adversarial training algorithm as a means to address the vulnerability of deep learning models to adversarial attacks. While existing research extensively explores the theoretical aspects of the training loss in adversarial training algorithms, this paper takes a distinct approach by examining the algorithmic stability, which contributes to establishing an upper bound for generalization error. The investigation into stability involves analyzing both upper and lower bounds. The paper contends that the non-differentiability issue inherent in adversarial training algorithms leads to poorer algorithmic stability compared to their natural counterparts. To mitigate this challenge, the study proposes a noise injection method. By addressing the non-differentiability issue through noise injection, the training trajectory becomes more likely to avoid instances of non-differentiability, resulting in improved stability performance for adversarial training. The analysis also explores the relationship between algorithm stability and the numerical approximation error associated with adversarial attacks.

\subsection{Pre-Processing}
A line of adversarial defense mechanisms make use of some pre-processing on the input samples in order to ensure that the samples will not behave adversarially once they are subjected to the actual machine learning models. These techniques make use of Data Compression, various kinds of Data Transformations and Dimensionality Reduction. They are explained in details here.

\begin{figure}[!htbp]
\centering
\includegraphics[width=\columnwidth]{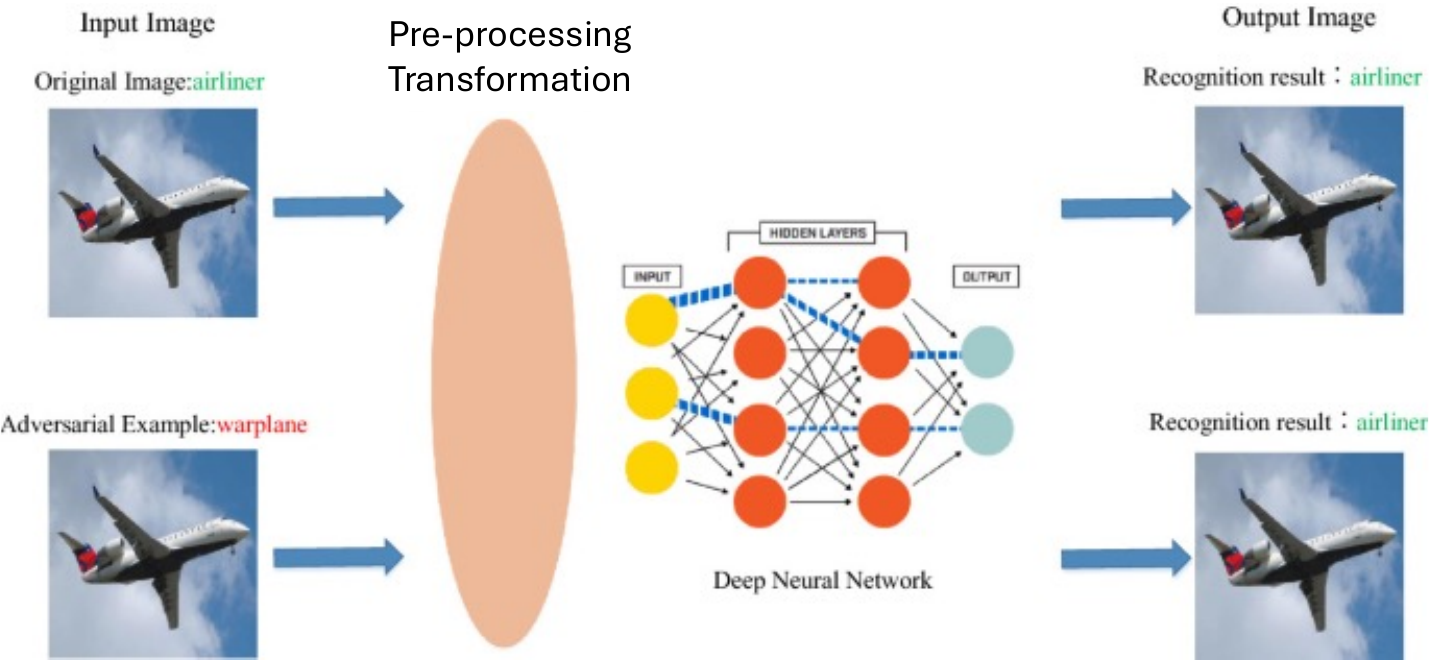}
\caption{These defense mechanisms make use of some pre-processing transformation on the input samples in order to ensure that the samples will not behave adversarially once they are subjected to the actual machine learning models.}
\label{preprocessing} 
\end{figure}

\subsubsection{Data Compression}
Deep neural networks (DNNs) have been found to be susceptible to adversarial examples, where imperceptible perturbations added to clean images can deceive well-trained networks. This paper introduces an end-to-end image compression model, named ComDefend \cite{comdefend}, as a defense mechanism against adversarial examples. ComDefend comprises a compression convolutional neural network (ComCNN) and a reconstruction convolutional neural network (ResCNN). The ComCNN preserves the structural information of the original image and eliminates adversarial perturbations, while the ResCNN reconstructs the original image with high quality. Essentially, ComDefend transforms adversarial images into their clean versions, which are then input to the trained classifier. Importantly, this method serves as a pre-processing module and does not modify the classifier's structure throughout the entire process. Hence, it can be seamlessly integrated with other model-specific defense models to collectively enhance the classifier's robustness. Experimental results conducted on MNIST, CIFAR10, and ImageNet demonstrate that the proposed method surpasses state-of-the-art defense techniques and consistently proves effective in safeguarding classifiers against adversarial attacks.

\subsubsection{Data Transformation}
Recent investigations have explored image compression-based strategies as a defense against adversarial attacks on deep neural networks (DNNs), which pose threats to their safe utilization. However, existing approaches predominantly rely on directly adjusting parameters such as compression rate to indiscriminately reduce image features. This approach lacks assurance in terms of both defense efficiency, measured by the accuracy of manipulated images, and the classification accuracy of benign images following the application of defense methods. To address these shortcomings, the authors propose a defensive compression framework based on JPEG, termed "feature distillation," \cite{feature_distillation} to effectively rectify adversarial examples without compromising the classification accuracy of benign data. The framework significantly enhances defense efficiency with minimal accuracy reduction through a two-step process: Firstly, it maximizes the filtering of malicious features in adversarial input perturbations by implementing defensive quantization in the frequency domain of JPEG compression or decompression, guided by a semi-analytical method. Secondly, it mitigates the distortions of benign features to restore classification accuracy through a DNN-oriented quantization refinement process.

\subsubsection{Dimension Reduction}
This paper investigates the impact of data dimensionality on adversarial examples, positing the hypothesis that generating adversarial examples is more straightforward in datasets with higher dimensions \cite{chattopadhyay}. The study delves into pertinent properties of high-dimensional spaces and provides empirical evidence through examinations on various models and datasets to validate the proposed hypothesis. The research specifically addresses the ease of generating adversarial examples, how this phenomenon behaves with different dimensionalities of input feature vectors, and the challenges associated with measuring adversarial perturbations at high dimensions using standard distance metrics such as $L_1$ and $L_2$ norms. The contributions of this work consist of two main elements: Theoretical Justification which offers the mathematical and statistical formulations pertinent to the classification of high-dimensional images to provide a rationale for the influence of dimensionality on adversarial example generation and Experimental Verification which presents the extensive empirical studies on image datasets with varying dimensions to gain insights into how dimensionality affects the generation of adversarial examples.

\begin{figure}[!htbp]
\centering
\includegraphics[width=\columnwidth]{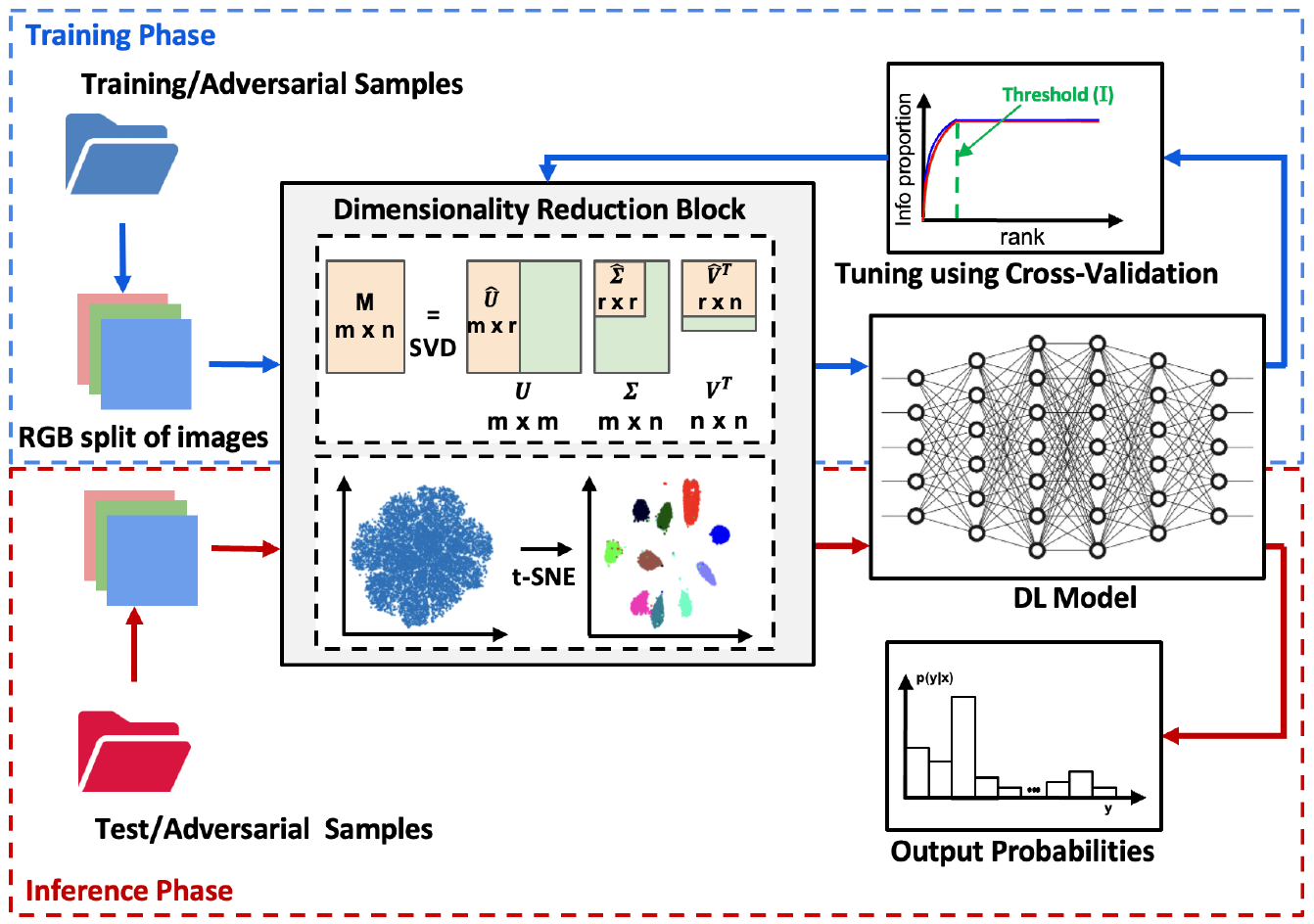}
\caption{Based on the observation that the high dimensional nature of the optimization landscape of neural networks contribute to adversarial vulnerability, Dimensionality Reduction is a useful technique in isolating adversarial noise present in samples and potentially eliminating the same.}
\label{DR} 
\end{figure}

\subsection{Defense of Patches}
An adversarial patch has the ability to manipulate image pixels within a defined region, leading to misclassification by the model. The localized nature of this attack has garnered considerable attention due to its practical applicability, enabling adversaries to execute physically-realizable attacks by attaching patches to target objects. There are adversarial defenses which are specifically designed to counter such attacks, as explained hereafter. 

\subsubsection{PatchGuard}
Recent defenses that offer provable robustness typically adopt the PatchGuard \cite{patchguard} framework, employing convolutional neural networks (CNNs) with small receptive fields and secure feature aggregation to ensure resilient model predictions. This paper introduces an extension of PatchGuard, named PatchGuard++, with the goal of provably detecting adversarial patch attacks to enhance both provable robust accuracy and clean accuracy. In PatchGuard++, the approach involves employing a CNN with small receptive fields for feature extraction, limiting the number of features affected by the adversarial patch. Subsequently, masks are applied in the feature space, and predictions are assessed on all possible masked feature maps. The final step involves extracting a pattern from all masked predictions to identify the presence of an adversarial patch attack. The evaluation of PatchGuard++ is conducted on ImageNette (a 10-class subset of ImageNet), ImageNet, and CIFAR-10. The results demonstrate a significant improvement in both provable robustness and clean performance achieved by PatchGuard++.

\subsubsection{Patch Cleanser}
\begin{figure}[!htbp]
\centering
\includegraphics[width=\columnwidth]{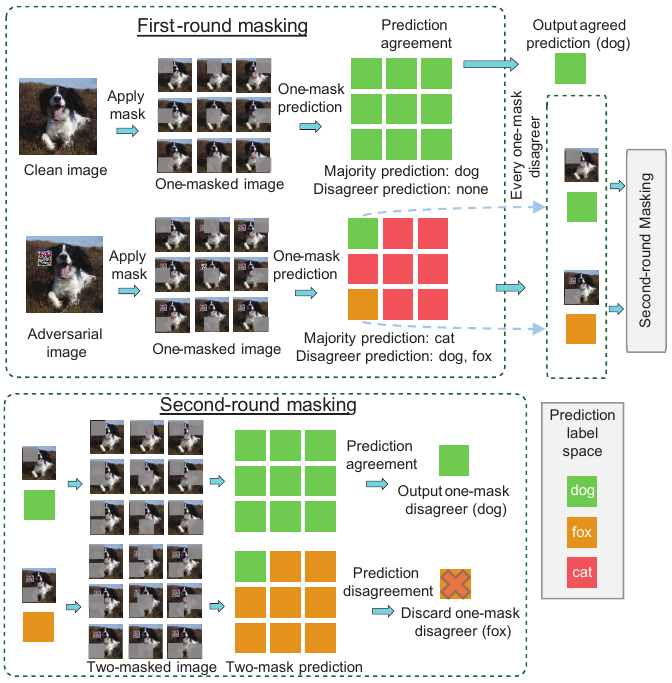}
\caption{Overview of double-masking defense: Masks are applied to input images to evaluate predictions. For clean images, predictions agree on the correct label. For adversarial images, disagreements are resolved by adding second-round masks. If two-mask predictions agree with a one-mask disagreer, its label is output; otherwise, it is discarded.}
\label{patchcleanser} 
\end{figure}

The objective of the adversarial patch attack in image classification models is to introduce adversarially crafted pixels into a defined image region (referred to as a patch), with the intention of causing misclassification by the model. This form of attack holds real-world implications, as it can be physically realized by printing and attaching the patch to the target object, posing a tangible threat to computer vision systems. To address this threat, the researchers propose PatchCleanser \cite{patchcleanser} as a certifiably robust defense mechanism against adversarial patches. The methodology is illustrated in Figure~\ref{patchcleanser}.

In the PatchCleanser approach, two rounds of pixel masking are applied to the input image, effectively neutralizing the impact of the adversarial patch. This operation in the image space ensures compatibility with any state-of-the-art image classifier, facilitating high accuracy. Importantly, the researchers can demonstrate that PatchCleanser consistently predicts the correct class labels on specific images, even against adaptive white-box attackers within the defined threat model, thus achieving certified robustness. The evaluation of PatchCleanser on datasets such as ImageNet, ImageNette, and CIFAR-10 reveals comparable clean accuracy to state-of-the-art classification models, along with a significant enhancement in certified robustness compared to previous works. Notably, PatchCleanser achieves an 83.9\% top-1 clean accuracy and a 62.1\% top-1 certified robust accuracy against a 2\%-pixel square patch positioned anywhere on the image for the 1000-class ImageNet dataset.

\subsubsection{Jedi}
\begin{figure}[!htbp]
\centering
\includegraphics[width=\columnwidth]{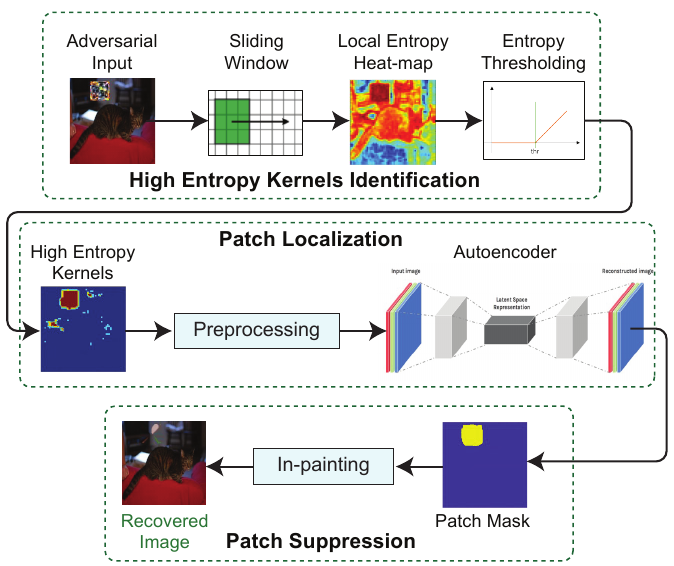}
\caption{Overview of the Jedi methodology: (1) High entropy kernels are identified using a sliding window and local entropy heat-maps. (2) Patch localization is refined using a sparse autoencoder. (3) Patch suppression is performed through in-painting to recover predictions.}
\label{Jedi} 
\end{figure}

In this research work, the authors introduce Jedi \cite{jedi}, a novel defense mechanism designed to effectively counter adversarial patches while demonstrating resilience against realistic patch attacks. Jedi addresses the challenge of patch localization through an information theory perspective, incorporating two innovative concepts: firstly, it enhances the identification of potential patch regions through entropy analysis, revealing that adversarial patches exhibit high entropy, even within naturalistic patches; secondly, it refines the localization of adversarial patches by employing an autoencoder capable of completing patch regions from high entropy kernels. The methodology is illustrated in Figure~\ref{Jedi}.

Jedi attains a high-precision adversarial patch localization, a crucial aspect for successfully repairing the images affected by such attacks. Notably, Jedi's reliance on input entropy analysis renders it model-agnostic, allowing its application to pre-trained off-the-shelf models without necessitating modifications to their training or inference processes. The evaluation of Jedi indicates its ability to detect, on average, 90\% of adversarial patches across various benchmarks and to recover up to 94\% of successful patch attacks. In comparison, other defense mechanisms such as LGS and Jujutsu achieve detection rates of 75\% and 65\%, respectively.

\subsubsection{ODDR}
%In this study, the researchers present Outlier Detection and Dimension Reduction (ODDR) \cite{oddr}, a comprehensive defense mechanism designed to proficiently counteract adversarial attacks based on patches. The proposed approach asserts that input features corresponding to adversarial patches, regardless of their naturalistic appearance, deviate from the inherent distribution of the remaining image sample and can be identified as outliers or anomalies. ODDR implements a three-stage pipeline—Fragmentation, Segregation, and Neutralization—providing a model-agnostic solution applicable to both image classification and object detection tasks.

%The Fragmentation stage dissects the samples into segments for the subsequent Segregation process, where outlier detection techniques pinpoint and isolate anomalous features linked to adversarial perturbations. The Neutralization stage employs dimension reduction methods on the outliers to alleviate the impact of adversarial perturbations while retaining essential information for the machine learning task.

%Extensive testing on benchmark datasets and state-of-the-art adversarial patches demonstrates the efficacy of ODDR. The results reveal robust accuracies that closely match or fall within a small range of clean accuracies (1\%-3\% for classification tasks), with only a marginal compromise of 1\%-2\% in performance on clean samples. This performance significantly outpaces other defense mechanisms.

This study introduces Outlier Detection and Dimension Reduction (ODDR) \cite{oddr}, a defense strategy designed to effectively mitigate adversarial patch-based attacks. The approach is based on the premise that input features associated with adversarial patches, even when visually natural, deviate from the underlying distribution of the rest of the image and can be detected as anomalies. ODDR employs a three-stage process—Fragmentation, Segregation, and Neutralization—offering a model-agnostic defense applicable to both image classification and object detection tasks.  

In the Fragmentation stage, image samples are divided into smaller segments, which are then analyzed in the Segregation stage to identify and isolate anomalous features indicative of adversarial perturbations using outlier detection techniques. The Neutralization stage applies dimension reduction to these outliers, reducing the influence of adversarial patches while preserving essential information necessary for accurate predictions.  

Extensive evaluations on benchmark datasets and advanced adversarial patches demonstrate ODDR’s effectiveness. Results indicate that model accuracy remains close to clean performance levels, with only a slight drop of 1\%-3\% for classification tasks and a minor 1\%-2\% reduction on clean samples, outperforming other existing defense methods.

\subsubsection{Local Gradients Smoothing}
%In response to recently introduced localized attacks such as Localized and Visible Adversarial Noise (LaVAN) and Adversarial patch, which present a novel challenge to the security of deep learning, a new method has been developed. This method addresses the addition of adversarial noise specifically within a defined region without affecting the salient objects in an image. The approach is motivated by the observation that these attacks introduce concentrated high-frequency changes at a specific image location. The proposed method, named Local Gradients Smoothing (LGS) \cite{lgs}, effectively estimates the location of noise in the gradient domain and transforms the high activation regions caused by adversarial noise in the image domain while minimizing the impact on salient objects crucial for correct classification. LGS achieves this by regularizing gradients in the estimated noisy region before inputting the image to the deep neural network (DNN) for inference. The effectiveness of the proposed method is demonstrated through comparisons with other defense techniques, including Digital Watermarking, JPEG compression, Total Variance Minimization (TVM), and Feature Squeezing, using the ImageNet dataset. Additionally, the robustness of the defense mechanism against Back Pass Differentiable Approximation (BPDA), a recently developed state-of-the-art attack, is systematically studied. In the context of localized adversarial attacks, LGS proves to be the most resistant to BPDA compared to other defense mechanisms.

To counter recently introduced localized attacks, such as Localized and Visible Adversarial Noise (LaVAN) and Adversarial Patch, which pose new challenges to deep learning security, a novel defense method has been developed. This approach specifically targets adversarial noise within a restricted region while preserving the integrity of salient objects in an image. The method is based on the observation that such attacks generate concentrated high-frequency alterations in a specific image area.  
The proposed technique, Local Gradients Smoothing (LGS) \cite{lgs}, identifies noisy regions in the gradient domain and modifies the corresponding high-activation areas in the image domain, minimizing disruption to key objects necessary for accurate classification. By regularizing gradients in the estimated noisy region before passing the image through a deep neural network (DNN), LGS effectively mitigates the impact of adversarial noise.  
The effectiveness of LGS is evaluated through comparisons with existing defense methods, including Digital Watermarking, JPEG compression, Total Variance Minimization (TVM), and Feature Squeezing, using the ImageNet dataset. Additionally, the method’s resilience against Back Pass Differentiable Approximation (BPDA), a cutting-edge adversarial attack, is systematically examined. Among various defenses against localized adversarial attacks, LGS demonstrates the highest resistance to BPDA.

\subsubsection{Jujutsu}
Jujutsu \cite{jujutsu} is a two-stage technique designed to detect and mitigate robust and universal adversarial patch attacks. The method begins by recognizing that adversarial patches are crafted as localized features exerting substantial influence on prediction outputs, maintaining dominance across various inputs. Jujutsu exploits this observation to achieve accurate attack detection with minimal false positives. Adversarial patches typically corrupt only a localized region of the input, leaving the majority of the input unperturbed. Leveraging this insight, Jujutsu utilizes generative adversarial networks (GANs) to perform localized attack recovery. This involves synthesizing the semantic contents of the input that are affected by the attacks and reconstructing a "clean" input conducive to correct predictions. The performance of Jujutsu is extensively evaluated on four diverse datasets, covering eight different deep neural network (DNN) models. The results demonstrate superior performance, significantly outperforming four existing defense mechanisms. Furthermore, Jujutsu undergoes evaluation against physical-world attacks and adaptive attacks, showcasing its effectiveness in diverse adversarial scenarios.

\subsection{Others}
The rest of the defense techniques, which have their own specific applications, are described here.

\subsubsection{Trapdoor Model}
In this research, the focus is on addressing the vulnerability of deep neural networks (DNNs) to adversarial attacks. Rather than attempting to patch weaknesses or increase the computational cost of generating adversarial examples, the authors propose a novel "honeypot" approach to safeguard DNN models. This approach deliberately introduces trapdoors—weaknesses in the classification manifold—to attract attackers seeking adversarial examples \cite{trapdoor}. The attackers' optimization algorithms are drawn toward these trapdoors, resulting in the production of attacks that resemble trapdoors in the feature space. The defense mechanism subsequently identifies attacks by comparing the neuron activation signatures of inputs to those of the trapdoors. The paper introduces and implements a trapdoor-enabled defense, providing analytical proof that trapdoors shape the computation of adversarial attacks, leading to feature representations similar to those of trapdoors. Experimental results demonstrate that models protected by trapdoors can accurately detect adversarial examples generated by state-of-the-art attacks, such as PGD, optimization-based CW, Elastic Net, and BPDA, with minimal impact on normal classification. These findings extend across various classification domains, including image, facial, and traffic-sign recognition. The study also presents significant results regarding the robustness of trapdoors against customized adaptive attacks or countermeasures.

\subsubsection{Defense against UAPs}
Adversarial training emerges as a robust defense against universal adversarial perturbation (UAP) by integrating corresponding adversarial samples into the training process. However, the utilization of adversarial samples in existing methods, such as UAP, introduces inevitable excessive perturbations associated with other categories, given its universal objective. Incorporating such samples into training leads to heightened erroneous predictions characterized by larger local positive curvature. This paper introduces a curvature-aware category adversarial training approach to mitigate excessive perturbations. The method introduces category-oriented adversarial masks synthesized with class-distinctive momentum. Additionally, the min-max optimization loops of adversarial training are split into two parallel processes to alleviate training costs. Experimental results conducted on CIFAR-10 and ImageNet demonstrate that the proposed method achieves superior defense accuracy against UAP with reduced training costs compared to state-of-the-art baselines.

\subsubsection{Feature Squeezing}
Prior research efforts aimed at defending against adversarial examples primarily concentrated on refining deep neural network (DNN) models, with limited success or requiring computationally expensive processes. The authors propose a novel strategy called feature squeezing to bolster DNN models by effectively detecting adversarial examples. Feature squeezing \cite{feature_squeezing} achieves this by narrowing down the search space available to adversaries, consolidating samples that correspond to various feature vectors in the original space into a single sample. By comparing a DNN model's predictions on the original input with those on squeezed inputs, feature squeezing demonstrates high accuracy in detecting adversarial examples with minimal false positives. This paper delves into two feature squeezing methods: reducing the color bit depth of each pixel and spatial smoothing. These straightforward strategies are cost-effective and can be employed in conjunction with other defense mechanisms. Furthermore, they can be combined within a joint detection framework to attain high detection rates against state-of-the-art adversarial attacks.

\subsubsection{Defense-GAN}
In recent years, the adoption of deep neural network approaches for various machine learning tasks, including classification, has become widespread. However, these models have demonstrated vulnerability to adversarial perturbations, where carefully crafted small changes can lead to misclassification of legitimate images. To address this challenge, the researchers introduce Defense-GAN, a novel framework that harnesses the expressive capability of generative models to protect deep neural networks from such attacks.

Defense-GAN \cite{defense_gan} is trained to model the distribution of unperturbed images. During inference, it identifies a close output to a given image that lacks adversarial changes and subsequently feeds this output to the classifier. Notably, the proposed method is compatible with any classification model, and it does not require modification of the classifier's structure or training procedure. Additionally, Defense-GAN serves as a defense against various attack methods, as it does not assume knowledge of the process for generating adversarial examples. Empirical results demonstrate the consistent effectiveness of Defense-GAN against different attack methods, showcasing improvements over existing defense strategies.

\subsubsection{MagNet}
MagNet \cite{magnet} presents a framework designed to defend neural network classifiers against adversarial examples without altering the protected classifier or possessing knowledge of the adversarial example generation process. The MagNet framework incorporates one or more distinct detector networks and a reformer network. In contrast to prior methods, MagNet learns to distinguish between normal and adversarial examples by approximating the manifold of normal examples. Its substantial generalization power stems from not relying on any specific process for generating adversarial examples. Furthermore, MagNet has the capability to reconstruct adversarial examples by guiding them toward the learned manifold. This proves effective in aiding the correct classification of adversarial examples with minimal perturbations. The paper addresses the inherent challenges in defending against whitebox attacks and introduces a mechanism to counter graybox attacks. Drawing inspiration from the use of randomness in cryptography, the authors propose incorporating diversity to enhance MagNet's robustness. Empirical results demonstrate MagNet's effectiveness against advanced state-of-the-art attacks in both blackbox and graybox scenarios while maintaining a low false positive rate on normal examples.

\subsubsection{High-Level Representation Guided Denoiser}
In addressing the vulnerability of neural networks to adversarial examples, particularly concerning their application in security-sensitive systems, a defense mechanism named high-level representation guided denoiser (HGD) \cite{hgd} is proposed for image classification. The conventional denoiser is susceptible to the error amplification effect, where slight residual adversarial noise is progressively magnified, leading to incorrect classifications. HGD tackles this issue by utilizing a loss function defined as the difference between the target model's outputs activated by the clean image and the denoised image. In comparison to the state-of-the-art defending method, ensemble adversarial training, which is effective on large images, HGD offers three notable advantages. Firstly, with HGD as a defense, the target model demonstrates enhanced robustness against both white-box and black-box adversarial attacks. Secondly, HGD can be trained on a small subset of images and exhibits strong generalization to other images and unseen classes. Thirdly, HGD is transferable and can defend models other than the one guiding its training. In the NIPS competition on defense against adversarial attacks, the HGD solution secured the first place, outperforming other models by a significant margin.

Other popular defense techniques for adversarial attacks include Input transformation-based
Sit: Stochastic input transformation to defend against adversarial attacks on deep neural networks \cite{guesmi2021sit}; different variants of Adversarial training like 
Room: Adversarial machine learning attacks under real-time constraints \cite{guesmi2022room} and
Exploring the Interplay of Interpretability and Robustness in Deep Neural Networks: A Saliency-guided Approach \cite{guesmi2024exploring}; and approximate hardware-based techniques like 
Defensive approximation: securing CNNs using approximate computing \cite{guesmi2021defensive} and  Defending with errors: Approximate computing for robustness of deep neural networks \cite{guesmi2022defending}. 

\section{Defenses in Object Detection}
\begin{figure}[!htbp]
\centering
\caption{Organization of different approaches for defenses against attacks on object detection tasks in vision based systems.}
\includegraphics[width=\columnwidth]{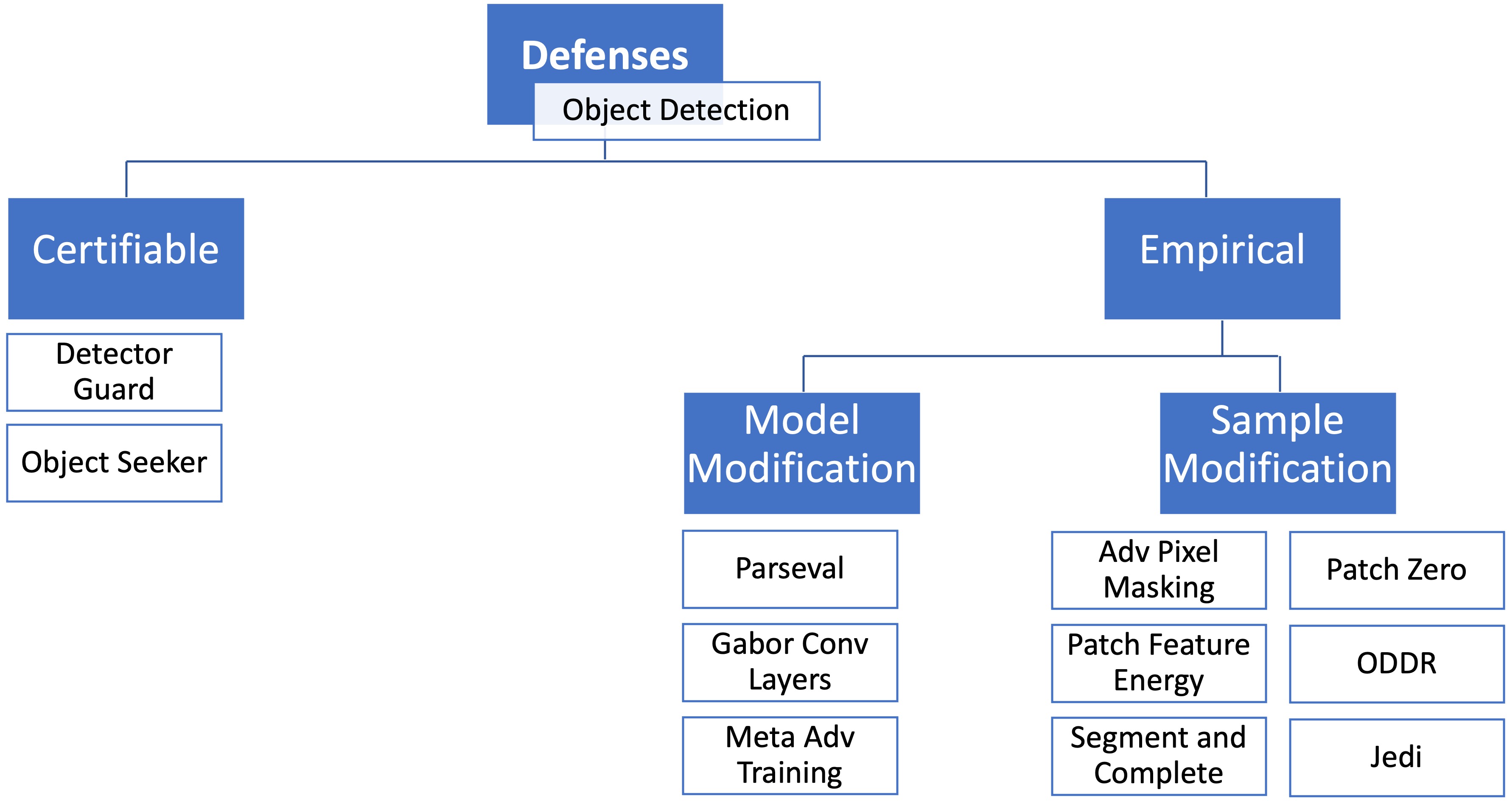}
\label{detection_image}
\end{figure}

In the face of an ever-evolving landscape of potent adversarial attacks on object detection tasks on video streams and dash-cam feeds of cars etc., there is a pressing need to identify and categorize corresponding defense strategies aimed at mitigating these threats. It is crucial to acknowledge that specific defenses exhibit robustness against particular attacks, and a universal solution for all challenges remains elusive. Consequently, we have systematically compiled the existing literature, delineating the applicability of these defenses and assessing their robustness against specific attacks in a Table \ref{detection_table_patch} and Table \ref{detection_table_others}. Furthermore, to enhance clarity on similar approaches, we have categorized and grouped these defense mechanisms, as depicted in the accompanying Figure \ref{detection_image}. The subsequent sections provide a brief introduction to each group along with their unique characteristics, followed by detailed explanations of individual methods.

\begin{table*}[!htbp]
\caption{Adversarial defenses in object detection tasks for patch based attacks.}
\label{detection_table_patch}
\centering
\begin{tabular}{llll}
\hline
\multirow{2}{*}{\textbf{Defense}} & \multirow{2}{*}{\textbf{Attacks Covered}} & \multirow{2}{*}{\textbf{Datasets}} & \multirow{2}{*}{\textbf{Best Performance}}  \\
 &  &  &  \\ \hline
\multirow{2}{*}{\begin{tabular}[c]{@{}l@{}}Meta Adversarial\\ Training \cite{meta}\end{tabular}} & \multirow{2}{*}{Patch Attack} & \multirow{2}{*}{Bosch Small Traffic Lights} & \multirow{2}{*}{47\% (Bosch Small Traffic Lights)}  \\
 &  &  &  \\
\multirow{2}{*}{\begin{tabular}[c]{@{}l@{}}Adversarial Pixel\\ Masking \cite{pixel_masking}\end{tabular}} & \multirow{2}{*}{Patch Attack} & \multirow{2}{*}{COCO, INRIA} & \multirow{2}{*}{92.2\% (INRIA)} \\
 &  &  &  \\
\multirow{2}{*}{\begin{tabular}[c]{@{}l@{}}Patch Feature\\ Energy \cite{patch_energy}\end{tabular}} & \multirow{2}{*}{\begin{tabular}[c]{@{}l@{}}Patches (LaVAN, Adv T-shirt \cite{adv_shirt}, \\ Adv-cloak \cite{adv_cloak}, Naturalistic)\end{tabular}} & \multirow{2}{*}{INRIA} & \multirow{2}{*}{91.4\% (INRIA)} \\
 &  &  &  \\
\multirow{2}{*}{\begin{tabular}[c]{@{}l@{}}Segment and\\ Complete \cite{segment}\end{tabular}} & \multirow{2}{*}{PGD, Dpatch, MIM} & \multirow{2}{*}{MS COCO, xVIEW} & \multirow{2}{*}{45.7\% (MS COCO)} \\
 &  &  &  \\
\multirow{2}{*}{\begin{tabular}[c]{@{}l@{}}Patch\\ Zero \cite{patchzero}\end{tabular}} & \multirow{2}{*}{Masked PGD} & \multirow{2}{*}{PASCAL VOC} & \multirow{2}{*}{66.1\% (PASCAL VOC)} \\
 &  &  &  \\
\multirow{2}{*}{Jedi \cite{jedi}} & \multirow{2}{*}{\begin{tabular}[c]{@{}l@{}}Patches (YOLO AP,\\ Naturalistic)\end{tabular}} & \multirow{2}{*}{INRIA, CASIA} & \multirow{2}{*}{88.2\% (CASIA)} \\
 &  &  &  \\
\multirow{2}{*}{\begin{tabular}[c]{@{}l@{}}Outlier Detection \&\\ Dimension Reduction \cite{oddr}\end{tabular}} & \multirow{2}{*}{\begin{tabular}[c]{@{}l@{}}Patches (YOLO AP, \\ Naturalistic)\end{tabular}} & \multirow{2}{*}{INRIA, CASIA} & \multirow{2}{*}{93.5\% (CASIA)} \\
 &  &  &  \\ \hline
\end{tabular}
\end{table*}

% \usepackage{multirow}
\begin{table*}[!htbp]
\caption{Adversarial defenses in object detection tasks for miscellaneous attack types.}
\label{detection_table_others}
\centering
\begin{tabular}{llll}
\hline
\multirow{2}{*}{\textbf{Defense}} & \multirow{2}{*}{\textbf{Attacks Covered}} & \multirow{2}{*}{\textbf{Datasets}} & \multirow{2}{*}{\textbf{Best Performance}}  \\
 &  &  &  \\ \hline
\multirow{2}{*}{\begin{tabular}[c]{@{}l@{}}Detector\\ Guard \cite{detectorguard}\end{tabular}} & \multirow{2}{*}{Adv T-shirt \cite{adv_shirt}, Adv-cloak \cite{adv_cloak}} & \multirow{2}{*}{\begin{tabular}[c]{@{}l@{}}PASCAL VOC, MS COCO, \\ KITTI\end{tabular}} & \multirow{2}{*}{32\% (KITTI)} \\
 &  &  &  \\
\multirow{2}{*}{\begin{tabular}[c]{@{}l@{}}Object\\ Seeker \cite{objectseeker}\end{tabular}} & \multirow{2}{*}{Adv T-shirt \cite{adv_shirt}, Adv-cloak \cite{adv_cloak}} & \multirow{2}{*}{PASCAL VOC, MS COCO} & \multirow{2}{*}{68.1\% (PASCAL VOC)} \\
 &  &  &  \\
\multirow{2}{*}{\begin{tabular}[c]{@{}l@{}}Parseval\\ Networks \cite{parseval}\end{tabular}} & \multirow{2}{*}{FGSM} & \multirow{2}{*}{SVHN} & \multirow{2}{*}{89.9\% (SVHN)} \\
 &  &  &  \\
\multirow{2}{*}{\begin{tabular}[c]{@{}l@{}}Gabor Convolutional\\ Layers \cite{gabor}\end{tabular}} & \multirow{2}{*}{TOG, DAG, RAP, UEA} & \multirow{2}{*}{PASCAL VOC, MS COCO} & \multirow{2}{*}{43.2\% (TOG)} \\
 &  &  &  \\ \hline
\end{tabular}
\end{table*}

\begin{figure}[!htbp]
\centering
\includegraphics[width=\columnwidth]{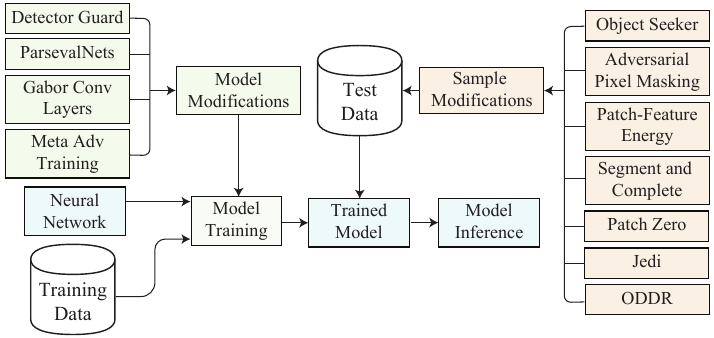}
\caption{Schematic representation of the integration of the defense techniques in different parts of a standard machine learning pipeline for object detection tasks.}
\label{survey_updated_2} 
\end{figure}
\vspace{-10pt}
\subsection{Certifiable Defenses}
The objective here is to create defenses that possess certifiable robustness. For a given clean data point (x, y), the defended model should consistently yield accurate predictions for any adversarial example falling within the defined threat model, meaning $D(x) = D(x) = y, \forall x \in A(x)$. The goal includes the development of a robustness certification procedure capable of verifying if the model's robustness can be certified. This certification process must consider all potential attackers within the specified threat model A, who may possess complete knowledge of the defense and unrestricted access to model parameters. Importantly, this certification establishes a provable lower bound for model robustness against adaptive attacks, offering a notable advancement over conventional empirical defenses susceptible to adaptive attackers.

\subsubsection{Detector Guard}
\begin{figure}[!htbp]
\centering
\includegraphics[width=0.9\columnwidth]{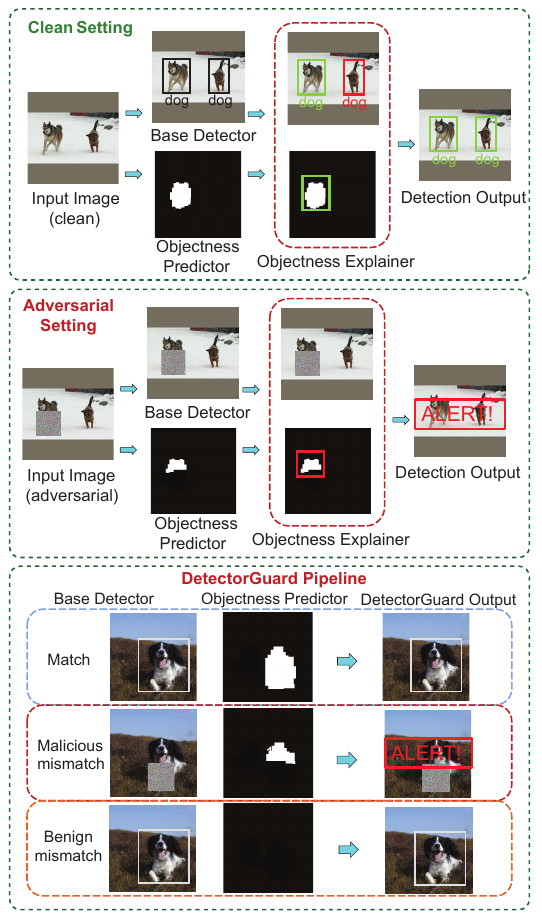}
\caption{DetectorGuard Overview: The Base Detector predicts bounding boxes, while the Objectness Predictor generates robust objectness maps. The Objectness Explainer matches predictions and identifies mismatches. On clean images, benign mismatches are accepted, and Base Detector outputs are used. In adversarial settings, unexplained objectness triggers attack alerts, ensuring robustness while maintaining clean performance.}
\label{Detectorguard} 
\end{figure}

In this research paper, DetectorGuard \cite{detectorguard} is introduced as the inaugural comprehensive framework designed for constructing object detectors that are provably robust against localized patch hiding attacks. Inspired by recent advancements in robust image classification, the study poses the question of whether robust image classifiers can be adapted for creating robust object detectors. Recognizing the task differences between the two, the paper addresses the challenges that arise in adapting an object detector from a robust image classifier. To overcome these challenges and develop a high-performance robust object detector, the proposed approach involves an objectness explaining strategy. This strategy entails adapting a robust image classifier to predict objectness (the probability of an object being present) for each image location. The predictions are then explained using bounding boxes generated by a conventional object detector. If all objectness values are well-explained, the output is based on the predictions of the conventional object detector; otherwise, an attack alert is issued. Significantly, the objectness explaining strategy provides provable robustness at no additional cost: 1) in the adversarial setting, the research formally establishes the end-to-end robustness of DetectorGuard on certified objects against any patch hiding attacker within the defined threat model; 2) in the clean setting, DetectorGuard exhibits almost identical performance to state-of-the-art object detectors. Evaluation results on the PASCAL VOC, MS COCO, and KITTI datasets illustrate that DetectorGuard achieves the first provable robustness against localized patch hiding attacks, with a negligible impact (< 1\%) on clean performance. The defense methodology is illustrated in Figure~\ref{Detectorguard}.

\subsubsection{Object Seeker}
In the context of security-critical systems like autonomous vehicles where object detectors are extensively utilized, susceptibility to patch hiding attacks has been identified. These attacks involve using a physically-realizable adversarial patch to cause the object detector to overlook the detection of specific objects, thereby compromising the functionality of object detection applications. The proposed solution, ObjectSeeker \cite{objectseeker}, aims to provide certifiably robust object detection against patch hiding attacks.

ObjectSeeker introduces a key insight known as patch-agnostic masking. This strategy involves the objective of masking out the entire adversarial patch without requiring knowledge of its shape, size, or location. Through this masking operation, the adversarial effect is neutralized, enabling any standard object detector to accurately detect objects within the masked images. Importantly, the robustness of ObjectSeeker can be formally evaluated in a certifiable manner. The authors have developed a certification procedure to determine if ObjectSeeker can detect specific objects against any white-box adaptive attack within the defined threat model, achieving certifiable robustness. Experimental results showcase a substantial improvement in certifiable robustness (10\%-40\% absolute and 2-6× relative) over prior works, while maintaining high clean performance (1\% drop compared to undefended models).

\subsection{Empirical - Model Modification}
Defenses against adversarial attacks in object detection tasks frequently encompass alterations to the model architecture or adjustments in the training process. These modifications and strategies are implemented to fortify the robustness of object detection models, aiming to ensure more dependable and secure performance in real-world scenarios. The specifics of each of these techniques are outlined below.

\subsubsection{Parseval}
Parseval networks \cite{parseval} have been introduced as a defense mechanism against adversarial attacks. This approach leverages the Lipschitz constant, based on the idea that a neural network, viewed as a composition of functions, can be made more resilient to small input perturbations by maintaining a low Lipschitz constant for these functions. The method achieves this by regulating the positional norm of the network’s weight matrices and parameterizing them using hard Parseval frameworks, leading to the development of what is termed Parseval networks.

\subsubsection{Gabor Convolutional Layer}
A recently proposed method enhances the robustness of object detectors against adversarial attacks by incorporating Gabor convolution layers \cite{gabor}. This approach first decomposes images into their RGB channels before processing them through a Gabor filter bank. Given their strong capability in extracting low-level image features, Gabor filters contribute to increased network resilience at this stage. The study demonstrates significant improvements in the performance of object detection models when exposed to adversarially manipulated images. Five robust object detection models are introduced and evaluated against various adversarial attacks. The proposed method achieves up to a 50\% enhancement in the performance of object detectors under adversarial conditions.

\subsubsection{Meta Adversarial Training}
Adversarial training stands out as the most potent defense mechanism against image-dependent adversarial attacks. Yet, customizing adversarial training for universal patches proves to be computationally demanding, given that the optimal universal patch is contingent on the model weights, which undergo changes during training. To address this hurdle, a novel approach named Meta Adversarial Training (MAT) \cite{meta} is introduced, seamlessly integrating adversarial training with meta-learning. MAT successfully tackles this challenge by meta-learning universal patches concurrently with model training. Notably, MAT demands minimal additional computational resources while dynamically adapting an extensive array of patches to the evolving model. The implementation of MAT significantly enhances robustness against universal patch attacks, particularly in the domains of traffic-light detection tasks.

\subsection{Empirical - Sample Modification}
Several adversarial defense mechanisms employ pre-processing techniques on input samples to prevent them from exhibiting adversarial behavior when presented to machine learning models. This involves applying certain transformations or modifications to input data before feeding it into the models, aiming to mitigate the susceptibility of the models to adversarial attacks. The goal is to pre-process the input in a way that enhances the model's ability to correctly classify and resist adversarial perturbations. Some techniques are explained here.

\subsubsection{Adversarial Pixel Masking}
In the realm of object detection relying on pre-trained deep neural networks (DNNs), notable advancements in performance have been achieved, enabling various applications. However, these DNN-based object detectors have demonstrated vulnerability to physical adversarial attacks. Despite recent efforts to develop defenses against such attacks, existing methods either rely on strong assumptions or exhibit reduced effectiveness when applied to pre-trained object detectors.

This paper introduces adversarial pixel masking (APM) \cite{pixel_masking} as a defense mechanism tailored for physical attacks, specifically designed for pre-trained object detectors. APM operates without necessitating assumptions beyond the "patch-like" nature of a physical attack and is compatible with different pre-trained object detectors featuring diverse architectures and weights. This adaptability positions APM as a practical solution across numerous applications. Extensive experiments validate the effectiveness of APM, demonstrating a significant improvement in model robustness without a substantial degradation of clean performance.

\subsubsection{Patch-Feature Energy}
In the context of security-critical systems, such as autonomous vehicles, the vulnerability of object detection to adversarial patch attacks has become evident. Current defense methods are constrained to dealing with localized noise patches by eliminating noisy regions in the input image. However, adversarial patches have evolved to adopt natural-looking patterns, eluding existing defenses. To tackle this challenge, the authors propose a defense method based on the innovative concept of "Adversarial Patch-Feature Energy" (APE) \cite{patch_energy}, which leverages common deep feature characteristics of an adversarial patch.

The proposed defense comprises APE-masking and APE-refinement, both designed to counter any adversarial patch reported in the literature. Comprehensive experiments demonstrate the effectiveness of the APE-based defense in achieving remarkable robustness against adversarial patches, both in digital environments and the physical world.

\subsubsection{Segment and Complete}
In addressing the critical yet insufficiently explored need for reliable defenses against patch attacks on object detectors, this paper introduces the Segment and Complete defense (SAC) \cite{segment} as a comprehensive framework. SAC aims to defend object detectors by detecting and eliminating adversarial patches. The approach involves training a patch segmenter that produces patch masks for pixel-level localization of adversarial patches. A self-adversarial training algorithm is proposed to enhance the robustness of the patch segmenter. Additionally, a robust shape completion algorithm is designed to ensure the removal of entire patches from images based on a specified Hamming distance between the segmenter's outputs and ground-truth patch masks.

Experiments conducted on COCO and xView datasets demonstrate the superior robustness of SAC, even under strong adaptive attacks, without compromising performance on clean images. SAC exhibits effective generalization to unseen patch shapes, attack budgets, and methods. The paper also introduces the APRICOT-Mask dataset, augmenting the APRICOT dataset with pixel-level annotations of adversarial patches. Results show that SAC significantly reduces the success rate of targeted physical patch attacks.

\subsubsection{Patch Zero}
%Adversarial patch attacks manipulate neural networks by introducing adversarial pixels within a localized region, offering high effectiveness across various tasks and physical realizability through attachment, such as using stickers on real-world objects. Despite the varied patterns of attack, adversarial patches typically exhibit a distinct texture and appearance compared to natural images. Capitalizing on this characteristic, the defense mechanism PatchZero \cite{patchzero} is introduced as a general pipeline against white-box adversarial patches, without requiring retraining of the downstream classifier or detector. The defense operates by detecting adversaries at the pixel level and "zeroing out" the patch region by repainting it with mean pixel values. Additionally, a two-stage adversarial training scheme is devised to enhance defense against stronger adaptive attacks. PatchZero demonstrates state-of-the-art defense performance in tasks such as image classification (ImageNet, RESISC45), object detection (PASCAL VOC), and video classification (UCF101), with minimal degradation in benign performance. Notably, PatchZero exhibits transferability to different patch shapes and attack types.

Adversarial patch attacks compromise neural networks by embedding adversarial pixels within a localized region, making them highly effective across various tasks and physically realizable, such as by attaching stickers to real-world objects. Despite the diversity in attack patterns, adversarial patches generally possess distinctive textures and appearances that differentiate them from natural images.  

Leveraging this property, PatchZero \cite{patchzero} is introduced as a general defense pipeline against white-box adversarial patches, eliminating the need for retraining the downstream classifier or detector. This method detects adversarial regions at the pixel level and neutralizes them by replacing the affected areas with mean pixel values. Additionally, a two-stage adversarial training approach is implemented to strengthen resilience against adaptive attacks.  

PatchZero achieves state-of-the-art defense performance across multiple tasks, including image classification (ImageNet, RESISC45), object detection (PASCAL VOC), and video classification (UCF101), while maintaining minimal impact on benign performance. Furthermore, the method demonstrates strong transferability to various patch shapes and attack strategies.

\begin{figure}[!htbp]
\centering
\includegraphics[width=0.9\columnwidth]{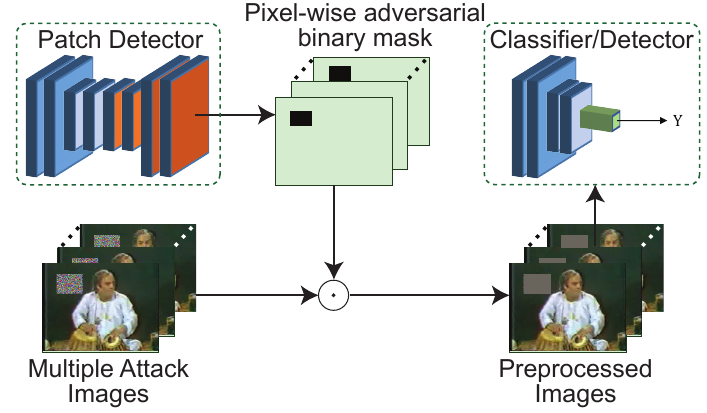}
    \caption{PatchZero Defense Pipeline: The patch detector predicts a binary mask \( M \) (black: adversarial, green: benign) for input images \( X \). Adversarial regions are zeroed out and filled with mean pixel values, producing \( X' \), which is passed to the downstream model for final prediction.}
\label{patchzero} 
\end{figure}

\subsubsection{Jedi}
In this study, the authors introduce Jedi \cite{jedi}, a novel defense mechanism designed to withstand realistic adversarial patch attacks. Jedi addresses the challenge of patch localization by adopting an information theory perspective and incorporating two innovative ideas. Firstly, it enhances the identification of potential patch regions through entropy analysis, demonstrating that adversarial patches exhibit high entropy even in naturalistic contexts. Secondly, Jedi improves the localization of adversarial patches by employing an autoencoder capable of completing patch regions from high entropy kernels. The defense achieves high-precision adversarial patch localization, a critical aspect for effectively restoring images. As Jedi relies on input entropy analysis, it is model-agnostic and can be applied to pre-trained off-the-shelf models without requiring modifications to their training or inference procedures. Evaluation across different benchmarks reveals that Jedi detects an average of 90\% of adversarial patches and successfully recovers up to 94\% of successful patch attacks, surpassing the performance of LGS and Jujutsu, which achieve 75\% and 65\%, respectively.

\subsubsection{ODDR}
This study introduces Outlier Detection and Dimension Reduction (ODDR) \cite{oddr}, a defense strategy designed to effectively mitigate adversarial patch-based attacks. The approach is based on the premise that input features associated with adversarial patches, even when visually natural, deviate from the underlying distribution of the rest of the image and can be detected as anomalies. ODDR employs a three-stage process—Fragmentation, Segregation, and Neutralization—offering a model-agnostic defense applicable to both image classification and object detection tasks.  

In the Fragmentation stage, image samples are divided into smaller segments, which are then analyzed in the Segregation stage to identify and isolate anomalous features indicative of adversarial perturbations using outlier detection techniques. The Neutralization stage applies dimension reduction to these outliers, reducing the influence of adversarial patches while preserving essential information necessary for accurate predictions.  

Extensive evaluations on benchmark datasets and advanced adversarial patches demonstrate ODDR’s effectiveness. Results indicate that model accuracy remains close to clean performance levels, with only a slight drop of 1\%-3\% for classification tasks and a minor 1\%-2\% reduction on clean samples, outperforming other existing defense methods.

\begin{figure}[!htbp]
\centering
\includegraphics[width=\columnwidth]{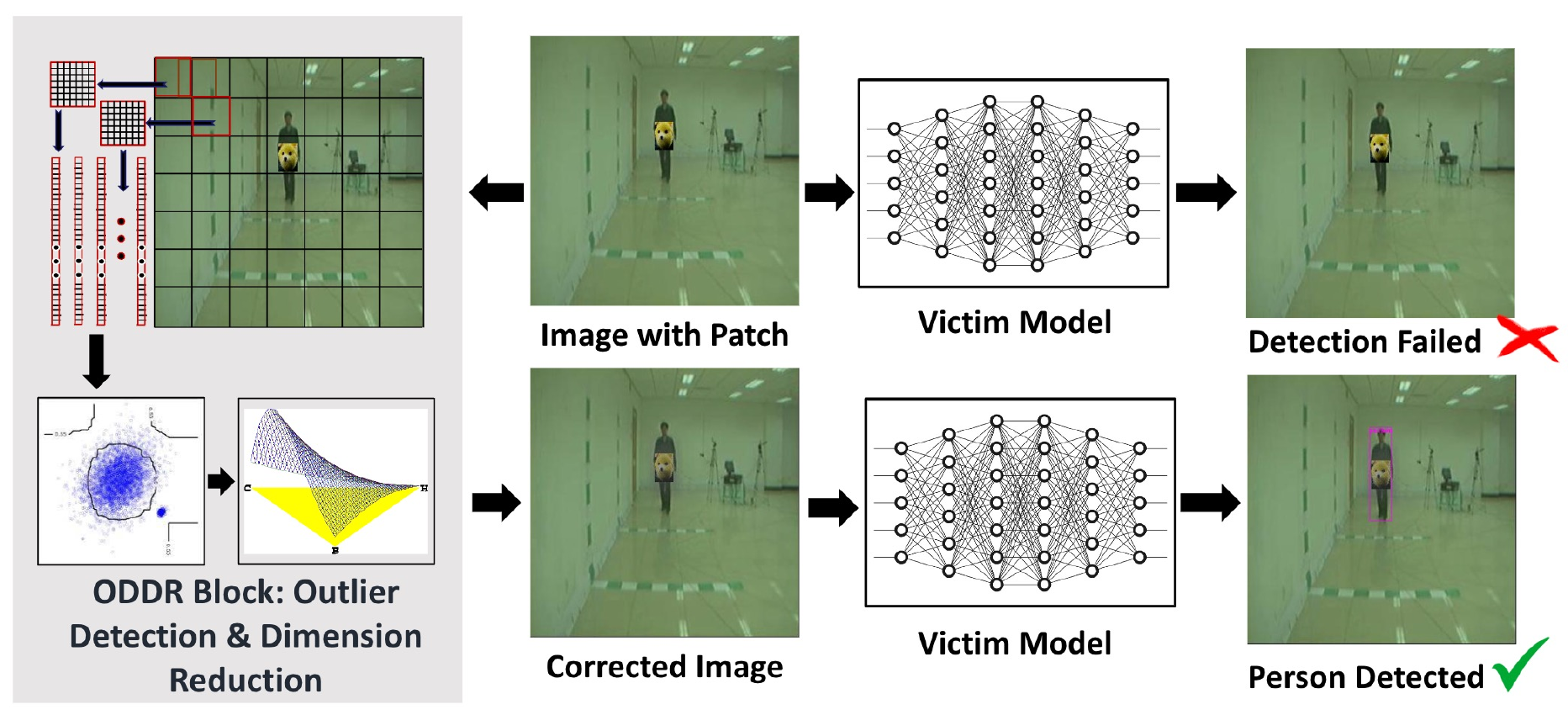}
\caption{It is observed that input features associated with adversarial patches, whether appearing naturalistic or otherwise, deviate from the inherent distribution of the remaining image sample and can be recognized as outliers or anomalies, making this strategy useful for practical defenses against patch-based adversarial attacks.}
\label{ODDR} 
\end{figure}

\section{Future Directions}
Future directions and research objectives in adversarial defenses for computer vision problems may include:
\begin{itemize}
    \item Robustness Across Domains: Enhancing defenses to ensure robustness across diverse domains and datasets, making models more adaptable to various real-world scenarios, investigating methods to improve the transferability of defenses, enabling models to generalize effectively across different architectures, tasks,
    \item Interpretability and Explainability: Developing defenses that provide better interpretability and explainability, allowing users to understand why a model makes certain predictions and facilitating trust in adversarial settings.
    \item Ensemble Approaches: Exploring ensemble-based defenses that combine multiple models or techniques to enhance robustness against a wide range of sophisticated and adaptive adversarial attacks.
    \item Dynamic Defenses: Designing defenses that can adapt dynamically to emerging attack strategies, ensuring continuous robustness against evolving adversarial techniques.
    \item Incorporating Domain Knowledge: Integrating domain-specific knowledge into defense mechanisms to tailor them for specific applications, such as medical imaging or autonomous driving. This may involve Human-in-the-Loop Defenses, which use the role of human intervention in the defense process, leveraging human perceptual abilities to identify and mitigate adversarial examples.
    \item Privacy-Preserving Defenses: Investigating defenses that not only protect against adversarial attacks but also preserve the privacy of sensitive information in the input data.
    \item Quantifiable Security Metrics: Developing standardized metrics to quantify the security and robustness of computer vision models against adversarial attacks, facilitating benchmarking and comparison.
\end{itemize}

These directions aim to advance the field of adversarial defenses in computer vision, making models more robust, interpretable, and resilient in practical applications.

\bibliographystyle{ieeetr}
\bibliography{bibliography}

\newpage
\newpage 
\pagebreak 
\clearpage
%%------------------------------
\begin{IEEEbiography}[{\includegraphics[width=1in,height=1.25in,clip,keepaspectratio]{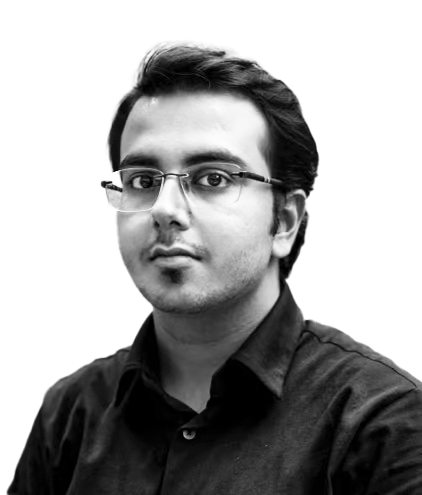}}]{Nandish Chattopadhyay} 
received the Doctoral degree in Computer Science and Engineering from Nanyang Technological University, Singapore, for his thesis on Robust Artificial Intelligence. He area of research includes security and privacy issues in machine learning, like adversarial attacks and defenses, backdoor attacks, watermarking neural networks for protecting IP rights and security of collaborative federated learning systems. Post PhD, he is currently a researcher at New York University (NYU) in Abu Dhabi, at the eBrain Lab at the Center for Cyber Security (CCS).
\end{IEEEbiography}

\begin{IEEEbiography}[{\includegraphics[width=1in,height=1.25in,clip,keepaspectratio]{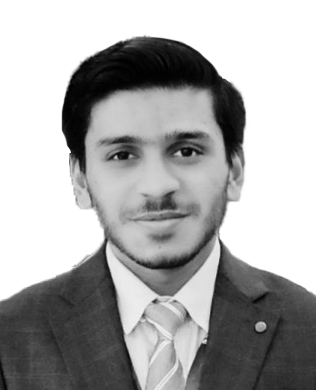}}]{Abdul Basit} 
received his B.S. degree in electrical engineering from the National University of Sciences and Technology (NUST), Pakistan, in 2021. He has been awarded the Rector's Gold Medal in B.S. EE. Since August 2023, he has been with New York University Abu Dhabi (NYUAD), where he is currently a Research Engineer in the eBrain Laboratory. His research interests include machine learning security, large language models, robotics systems, and autonomous driving.
\end{IEEEbiography}

%\begin{IEEEbiography}[{\includegraphics[width=1in,height=1.25in,clip,keepaspectratio]{Fig/Authors/guesmi.jpg}}]{Amira Guesmi} received the Engineer degree in Computer Science \& Electrical Engineering, from National School of Engineers of Sfax, Tunisia, in 2016, and the Ph.D. degree in Computer Systems Engineering from Polytechnic University Hauts-De-France, France, and the National School of Engineers of Sfax, Tunisia, in 2021. Afterwards, she worked as a Postdoctoral researcher at IEMN-DOAE Laboratory (CNRS-8520), Université Polytechnique Hauts-de-France. Dr. Guesmi is currently working as a research group leader in New York University (NYU) Abu Dhabi, UAE. Her research interests include AI safety, Machine Learning Security and Privacy, and Lifelong Learning.
%\end{IEEEbiography}

\begin{IEEEbiography}[{\includegraphics[width=1in,height=1.25in,clip,keepaspectratio]{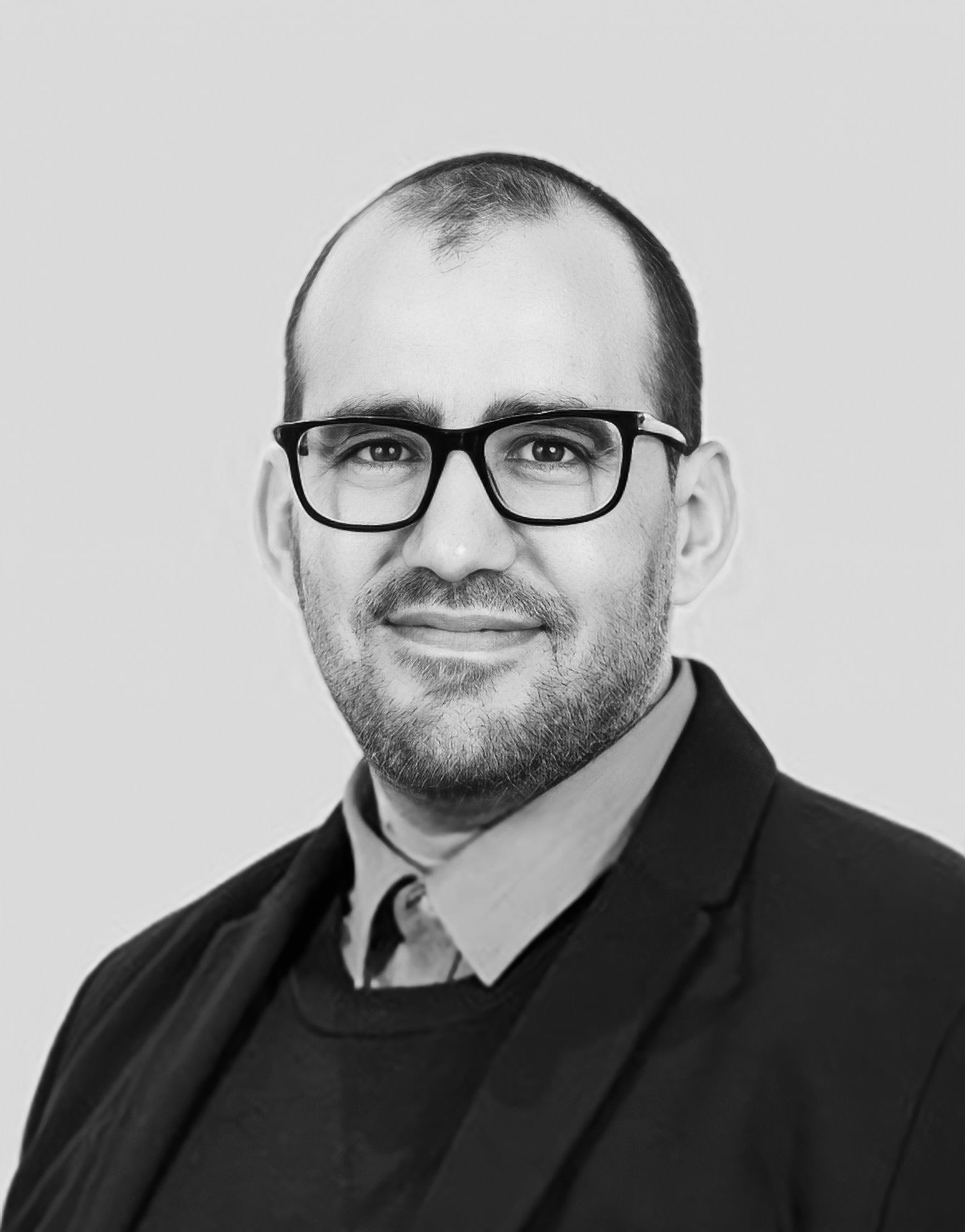}}]{Bassem Ouni} Dr./Eng. Bassem Ouni is currently a Lead Researcher at Technology Innovation Institute, Abu Dhabi,
United Arab Emirates. He received his Ph.D. degree in Computer Science from the University of Nice-Sophia Antipolis, Nice, France, in July 2013. Between October 2018 and January 2022, he held a Lead Researcher position in the French Atomic Energy Commission (CEA) within the LIST Institute, Paris, France, and an Associate Professor/Lecturer position at the University of Paris Saclay and ESME Sudria Engineering school, Paris, France. Prior to that, he worked as a Lead Researcher between 2017 and 2018 within the department of Electronics and Computer Science, University of Southampton, Southampton, United Kingdom. Before that, he occupied the position of a Research Scientist, between 2015 and 2016, at the Institute of Technology in Aeronautics, Space and Embedded Systems (IRT-AESE) located in Toulouse, France. From September 2013 to the end of 2014, he held a post-doctoral fellow position in Eurecom, Telecom ParisTech institute, Sophia Antipolis, France. Furthermore, he worked between 2009 and 2013 as a lecturer at the University of Nice Sophia Antipolis (Polytech Nice Engineering School and Faculty of
Sciences of Nice). Also, he was managing several industrial collaborations with ARM, Airbus Group Innovation, Rolls Royce, Thales Group, Continental, Actia Automotive Group, etc. He co-authored many publications (Book Chapters, Journals, and international conferences.). He is an IEEE senior member.
\end{IEEEbiography}
\vspace{-30pt}
\begin{IEEEbiography}[{\includegraphics[width=1in,height=1.25in,clip,keepaspectratio]{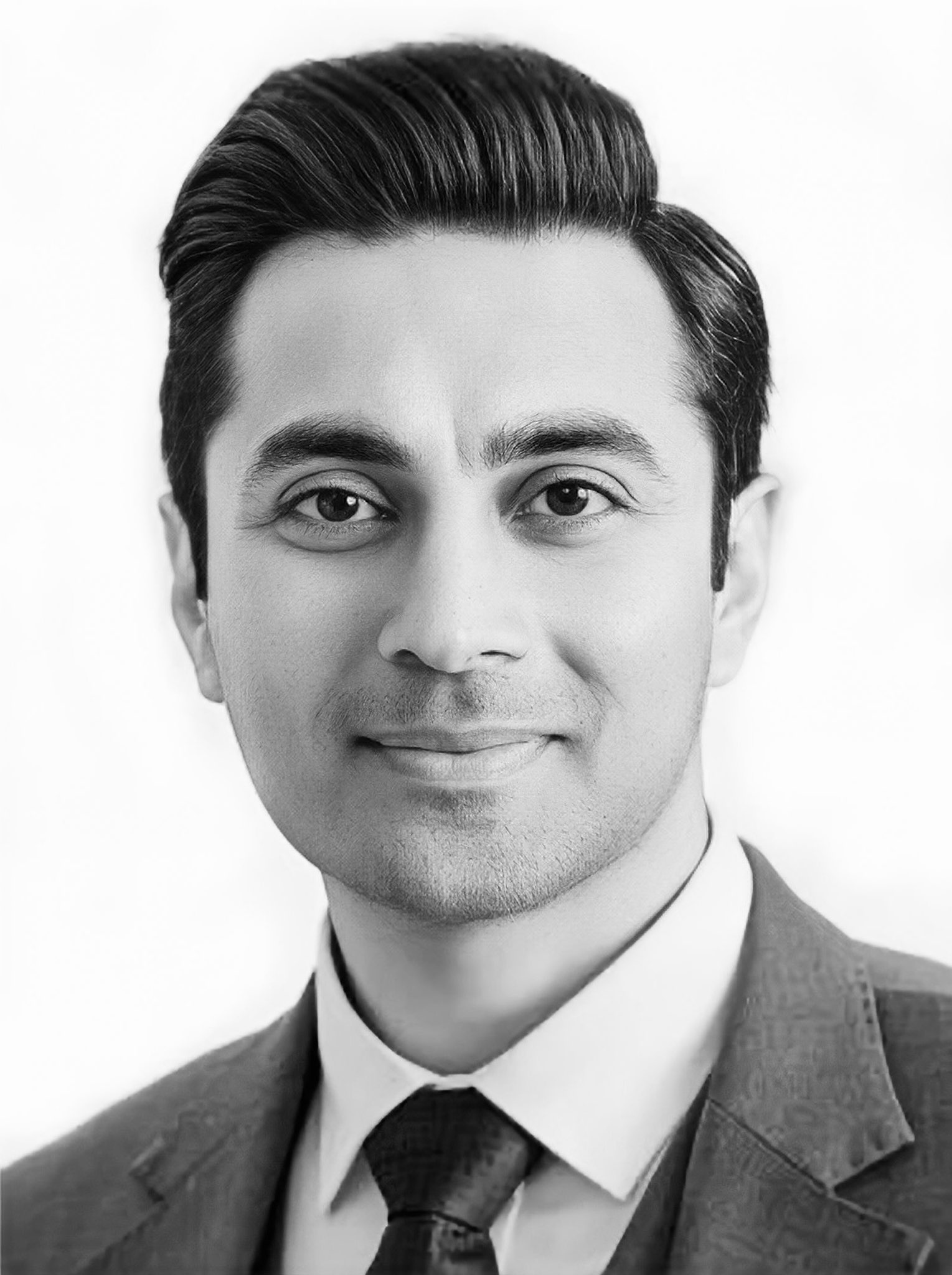}}]{Muhammad Shafique} (M’11 - SM’16) received the Ph.D. degree in computer science from the Karlsruhe Institute of Technology (KIT), Germany, in 2011. Afterwards, he established and led a highly recognized research group at KIT for several years as well as conducted impactful collaborative R\&D activities across the globe. In Oct.2016, he joined the Institute of Computer Engineering at the Faculty of Informatics, Technische Universität Wien (TU Wien), Vienna, Austria as a Full Professor of Computer Architecture and Robust, Energy-Efficient Technologies. Since Sep.2020, Dr. Shafique is with the New York University (NYU), where he is currently a Full Professor and the director of eBrain Lab at the NYU-Abu Dhabi in UAE, and a Global Network Professor at the Tandon School of Engineering, NYU-New York City in USA. He is also a Co-PI/Investigator in multiple NYUAD Centers, including Center of Artificial Intelligence and Robotics (CAIR), Center of Cyber Security (CCS), Center for InTeractIng urban nEtworkS (CITIES), and Center for Quantum and Topological Systems (CQTS).
His research interests are in AI \& machine learning hardware and system-level design, brain-inspired computing, machine learning security and privacy, quantum machine learning, cognitive autonomous systems, wearable healthcare, energy-efficient systems, robust computing, hardware security, emerging technologies, FPGAs, MPSoCs, and embedded systems. He has given several Keynotes, Invited Talks, and Tutorials, as well as organized many special sessions at premier venues. He has served as the PC Chair, General Chair, Track Chair, and PC member for several prestigious IEEE/ACM conferences. He holds one U.S. patent, and has (co-)authored 6 Books, 10+ Book Chapters, 350+ papers in premier journals and conferences, and 100+ archive articles. He received the 2015 ACM/SIGDA Outstanding New Faculty Award, the AI 2000 Chip Technology Most Influential Scholar Award in 2020 and 2022 the AI 2000 Chip Technology Most Influential Scholar Award in 2020, 2022, and 2023, the ASPIRE AARE Research Excellence Award in 2021, six gold medals, and several best paper awards and nominations at prestigious conferences. He is a senior member of the IEEE and IEEE Signal Processing Society (SPS), and a member of the ACM, SIGARCH, SIGDA, SIGBED, and HIPEAC.
\end{IEEEbiography}

\EOD
\end{document}